\IfFileExists{latexml.sty}{\RequirePackage{latexml}}{\newif\iflatexml}
\iflatexml
  \documentclass{article}
  \usepackage{authblk,graphicx,multirow,xcolor}
  \usepackage[numbers,sort&compress]{natbib}
  \usepackage{hyperref}
\else
  \documentclass[]{onethree}
\fi

\usepackage{booktabs}
\usepackage{tabularx}
\usepackage{colortbl}
\usepackage[toc,page,header]{appendix}
\usepackage{minitoc}
\usepackage{amsmath,amssymb}
\usepackage{bm}
\usepackage{wrapfig}
\usepackage{placeins}
\usepackage{float}
\usepackage[bottom]{footmisc}
\usepackage{threeparttable}
\usepackage{enumitem}
\usepackage{CJKutf8}
\iflatexml\else
  \usepackage{needspace}
\fi

\title{MNN-Pocket-0.3B Technical Report}

\iflatexml
  \author{MNN Team}
  \affil{Taobao and Tmall Group of Alibaba}
  \date{}
\else
  \author[1]{MNN Team}
  \affiliation[1]{Taobao and Tmall Group of Alibaba}
\fi

\abstract{
This technical report presents \textbf{MNN-Pocket-0.3B}, a compact 300M-parameter dense language model designed for efficient on-device deployment and supporting context lengths of up to 32K tokens. With an INT4 deployment footprint of only 173 MB and decoding throughput of up to 466 tokens/s on iPad Pro M5 (Metal), it can be seamlessly integrated into applications on edge devices such as smartphones and smart glasses.

MNN-Pocket-0.3B is developed through the co-design of its architecture, data, training pipeline, and inference system. Hardware-aware architecture selection and dependency-aware vocabulary pruning balance model capability with on-device memory and compute constraints. A four-stage pre-training pipeline covering approximately 4.1 trillion high-quality Chinese and English tokens equips the model with broad knowledge and reasoning capabilities. This is followed by a three-stage post-training pipeline comprising instruction fine-tuning, multi-teacher on-policy distillation, and quantization-aware training. Finally, hardware-specific optimizations in the MNN inference engine enable fast, memory-efficient inference and long-context processing.

Extensive evaluations demonstrate that MNN-Pocket-0.3B achieves frontier-level performance among models of comparable size, with particularly strong results in Chinese and English knowledge and reasoning, mathematics, coding, and instruction following. It consistently outperforms similarly sized open-source models and even surpasses several 0.5B-parameter models on instruction-following benchmarks.

We release MNN-Pocket-0.3B with open weights and ready-to-use MNN deployment packages. We hope it will provide a practical foundation for capable, responsive, and memory-efficient edge applications while advancing research and development in on-device language models.
\vspace{-0.1cm}
}

\iflatexml\else
\checkdata[Resources]{
  \href{https://github.com/alibaba/mnn}{\raisebox{-1ex}{\includegraphics[height=4.2ex]{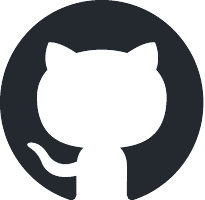}}~GitHub}\hspace{1em} \
  \href{https://huggingface.co/taobao-mnn/MNN-Pocket-0.3B}{\raisebox{-1ex}{\includegraphics[height=4.2ex]{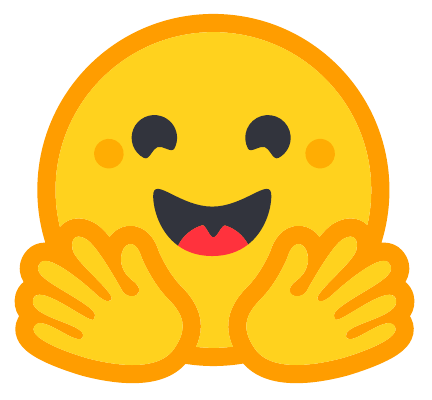}}~Hugging Face}\hspace{1em} \
  \href{https://modelscope.cn/models/MNN/MNN-Pocket-0.3B}{\raisebox{-1ex}{\includegraphics[height=4.2ex]{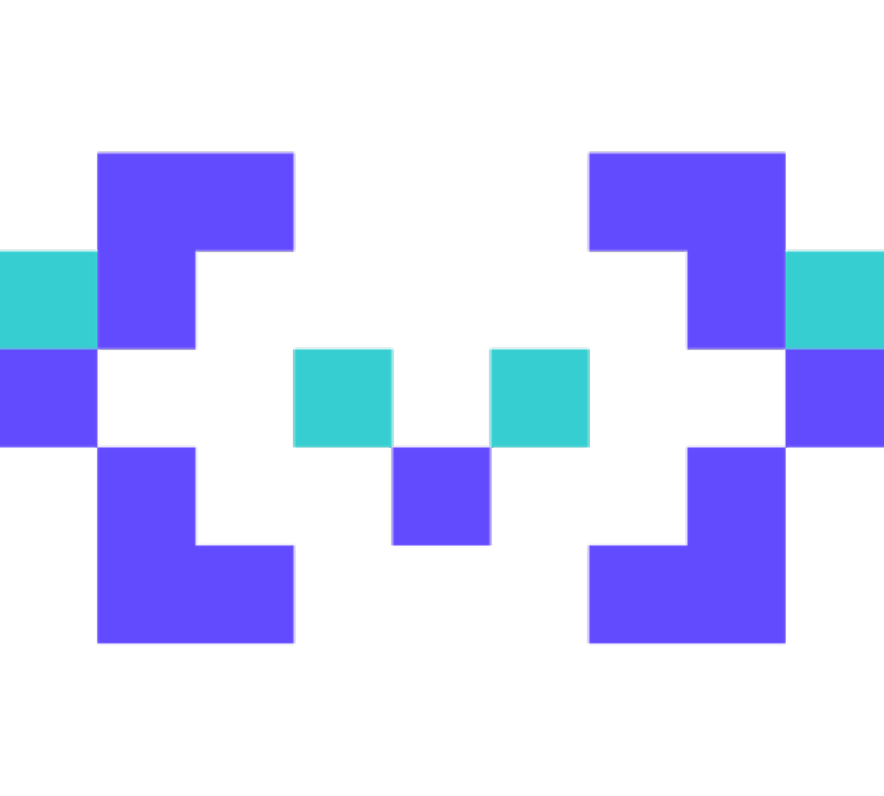}}~ModelScope}\hspace{1em} \
}
\fi

\begin{document}
\begin{CJK*}{UTF8}{gbsn}
\maketitle
\iflatexml
  \noindent\textbf{Resources.}\enspace
  \href{https://github.com/alibaba/mnn}{GitHub}\quad
  \href{https://huggingface.co/taobao-mnn/MNN-Pocket-0.3B}{Hugging Face}\quad
  \href{https://modelscope.cn/models/MNN/MNN-Pocket-0.3B}{ModelScope}\par
\else
  \vspace{-1.1cm}
\fi

\begin{figure}[!htbp]
  \centering
  \includegraphics[width=1.0\linewidth]{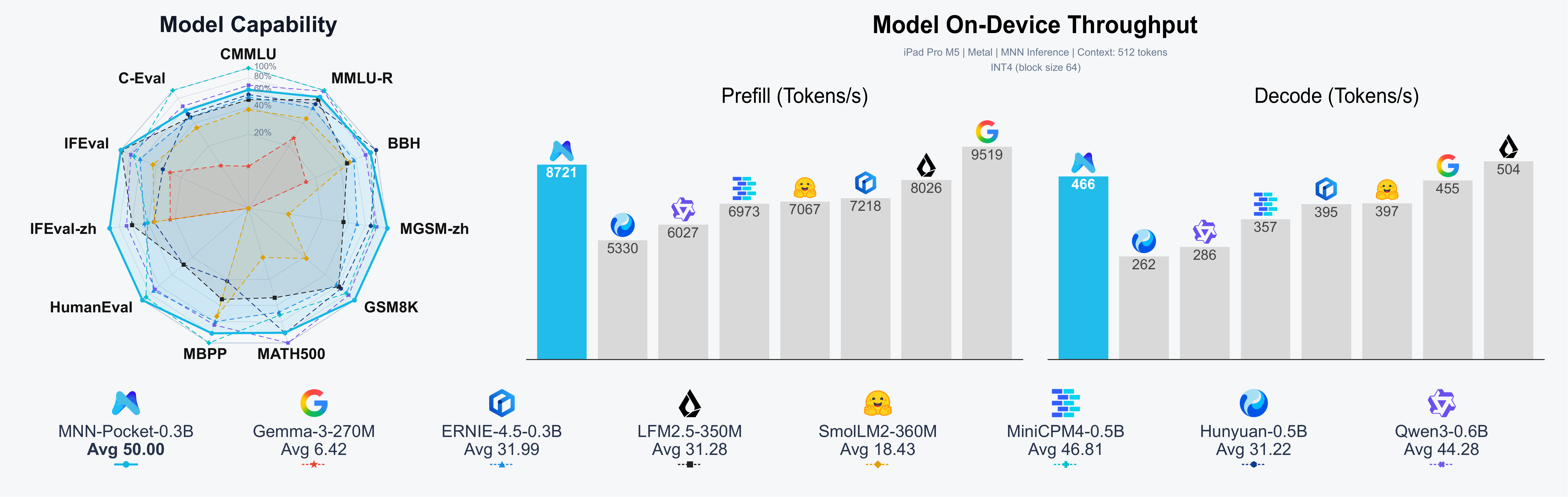}
  \caption{\textbf{MNN-Pocket-0.3B Model Overview.} }
  \label{fig:teaser}
\end{figure}

\newpage

\tableofcontents
\newpage


\section{Introduction}

On-device intelligence has already created substantial value in applications such as search and recommendation, short-form video, live streaming, and extended reality. In the era of large language models, a unified on-device model has the potential to consolidate multiple task-specific models while delivering low-latency responses, reducing cloud inference costs, protecting user privacy, supporting offline operation, and simplifying model maintenance and iteration.

Recent work has explored compact LLMs ranging from sub-billion to a few billion parameters. SmolLM~\cite{SmolLM}, LFM~\cite{LFM,LFM2.5}, and MiniCPM~\cite{MiniCPM} focus specifically on small language models, while major open model families—including Gemma~\cite{Gemma}, Qwen ~\cite{Qwen3}, Hunyuan~\cite{Hunyuan}, and ERNIE~\cite{ERNIE}—offer variants as small as 0.3–0.6B parameters. Despite this progress, deployment in large-scale consumer applications remains challenging. Such applications must support diverse mobile hardware under stringent resource budgets, often requiring 200–300 MB of resident memory and less than 300 MB of peak memory, while targeting generation speeds approaching 200 tokens/s. Existing models rarely balance memory footprint, inference speed, and task performance simultaneously. Addressing this gap requires an edge-first approach in which the model and inference system are jointly optimized across architecture, data, training, quantization, and runtime execution.

We therefore introduce MNN-Pocket-0.3B, a compact 300M-parameter dense language model explicitly designed for on-device deployment. It supports context lengths of up to 32K tokens, occupies only 173 MB with INT4 weights, and achieves decoding throughput of up to 466 tokens/s on iPad Pro M5 (Metal). These properties make MNN-Pocket-0.3B practical for integration into resource-constrained applications on smartphones, smart glasses, and other edge platforms.

MNN-Pocket-0.3B is built around four key innovations:
\begin{itemize}
    \item \textbf{Edge-first architecture–tokenizer co-design.} Guided by on-device measurements with MNN~\cite{jiang2020mnn,mnn-llm}, we select a 300M-parameter dense decoder-only architecture and refine its depth–width allocation to balance model quality, memory footprint, and decoding throughput. We complement this design with a compact 64K bilingual tokenizer derived through dependency-aware vocabulary pruning, balancing sequence compression against embedding memory and output-projection cost.
    \item \textbf{Data-centric, stage-aware pre-training.} We curate and synthesize Chinese, English, math and code data, grouping samples by domain, quality, knowledge density, difficulty, and style. Proxy runs and model-update interpolation screen candidate recipes, which are refined through direct training on data mixtures. The four-stage pipeline coordinates data mixtures and learning-rate schedules across approximately 4.1 trillion tokens to build knowledge and reasoning capabilities within a compact model.
    \item \textbf{Capability-centric, deployment-aware post-training.} A three-stage pipeline combines supervised fine-tuning with the SFT Loop Engine, multi-teacher on-policy distillation (MOPD), and quantization-aware training (QAT). MOPD routes student-generated trajectories to domain-specialized teachers, consolidating knowledge, instruction following, mathematics, and coding capabilities into a single compact model. Finally, Self-OPD with QAT aligns the quantization-aware student with its frozen BF16 counterpart to mitigate quantization-induced degradation for W4A16 deployment.
    \item \textbf{Model–runtime optimization for efficient edge deployment.} MNN-Pocket-0.3B is deployed with MNN through standardized model assets and a unified inference interface with hardware-specific backends. Phase-aware low-bit kernels, fused Transformer operators, and backend-specific attention and KV-cache implementations enable efficient INT4 inference, achieving 8,721 tokens/s in prefill and 466 tokens/s in autoregressive decoding on an iPad Pro M5 with Metal. Optional speculative decoding further accelerates generation, with gains depending on the device, backend, and workload.
\end{itemize}

The remainder of this report is organized as follows. Section 2 presents the architecture–tokenizer co-design of the MNN-Pocket-0.3B backbone. Sections 3 and 4 describe the pre-training and post-training pipelines, respectively. Section 5 details deployment with MNN-LLM, including runtime optimizations and evaluations of inference speed and memory efficiency. Section 6 reports comprehensive benchmark results across Chinese and English knowledge and reasoning, mathematics, coding, and instruction following. Finally, Section 7 discusses potential applications of MNN-Pocket-0.3B on edge platforms.

\newpage

\section{Architecture}


\subsection{Design Objectives}

On-device LLM design requires a different objective from frontier-scale models: instead of maximizing capability under abundant resources, we optimize jointly for model quality, training efficiency, inference latency, memory footprint, and deployment simplicity within tight resource budgets.

\begin{itemize}
    \item \textbf{Low-latency and memory-efficient inference.} For on-device deployment (e.g., smartphones and wearables), low latency and small memory footprint are primary requirements. This differs fundamentally from training, which is typically performed on multi-GPU clusters optimized for throughput and scale. As a result, an architecture that performs slightly better during training may still be unsuitable for deployment if it introduces irregular memory access, increases KV-cache overhead, or is poorly supported by optimized inference kernels. We prioritize designs that execute efficiently on practical inference engines and hardware backends.
    \item \textbf{High quality under tight parameter budgets.} Compared with hundred-billion-parameter models, models in the sub-billion-parameter regime are more sensitive to how parameters are allocated across attention, feed-forward layers, embeddings, and depth~\cite{pmlr-v235-liu24ce}. Our goal is therefore not merely to reduce parameter count, but to maximize capability under a limited parameter budget and deployment-friendly constraints.
    \item \textbf{Training efficiency and optimization stability.} For on-device LLMs, training on GPUs is intrinsically less efficient: reduced model and activation sizes often lead to lower memory occupancy and Model Flops Utilization (MFU), making it harder to fully utilize modern accelerators. The architecture should therefore support stable optimization, efficient hardware utilization, and predictable scaling under moderate batch sizes and limited token budgets.
    \item \textbf{Simplicity and extensibility.} Although modern Transformer variants~\cite{cao2026qwen3,kimik3} can improve efficiency in specific scenarios, such as long-context modeling, they often introduce greater training complexity and weaker compatibility with on-device deployment stacks. Our goal is to adopt a design that is easy to train, efficient to deploy on edge hardware, and extensible to future model variants.
\end{itemize}
Taken together, our goal is to identify the most effective on-device LLM configuration by jointly balancing quality, efficiency, latency, and flexibility.

\subsection{Scaling On-Device Dense Model}

We adopt an architecture scaled down from the dense Qwen3 model~\cite{Qwen3} as the backbone of MNN-Pocket-0.3B.
While mixture-of-experts (MoE) architectures can increase model capacity without proportionally increasing the number of active parameters, they require a substantially larger total parameter footprint and introduce routing and non-contiguous expert access~\cite{chen2026mobilemoe}. These costs are unfavorable for batch-one on-device decoding, where memory capacity, bandwidth, and access locality are more critical than peak compute throughput.

\begin{figure}[!t]
    \centering
    \includegraphics[width=\textwidth]{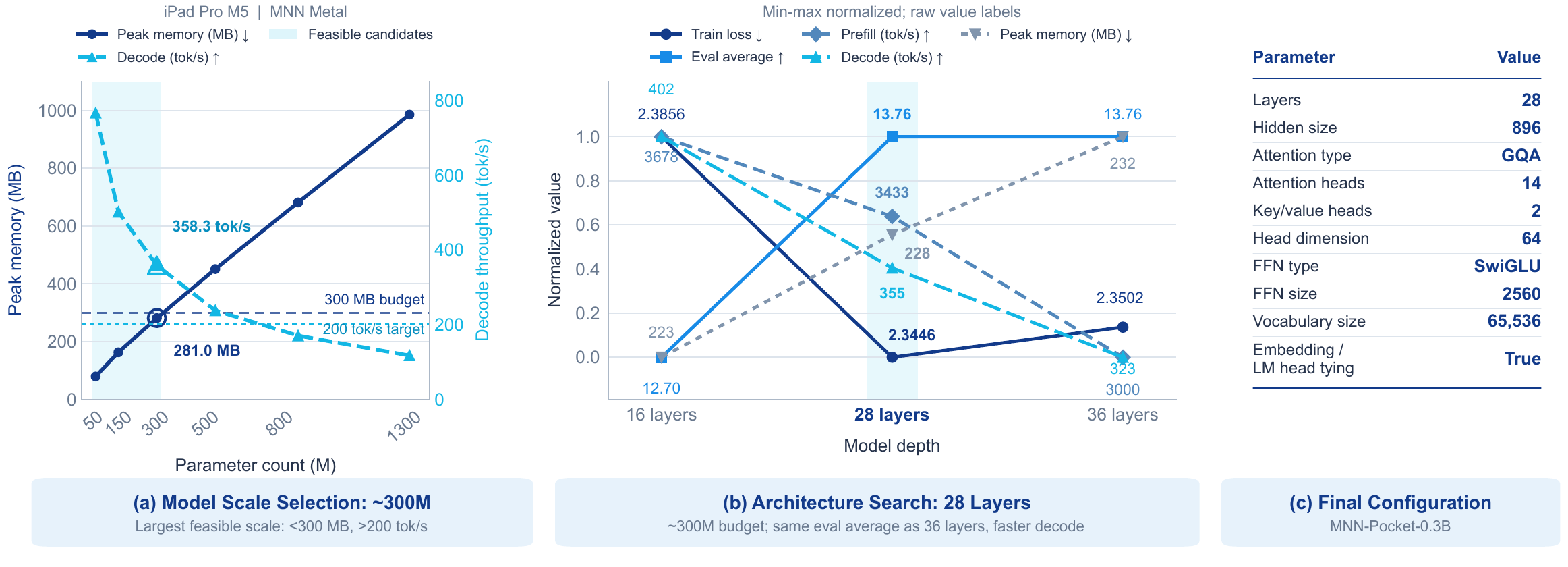}
    \caption{
        \textbf{Hardware-aware architecture selection.}
        (a) Model-scale selection on iPad Pro M5 using the MNN Metal backend. The approximately 300M-parameter model is the largest tested candidate with peak memory below 300 MB and decoding throughput above 200 tokens/s; shading marks the feasible tested candidates.
        (b) Architecture search at a comparable 300M-parameter budget. The 28-layer model achieves the same evaluation average as the 36-layer model, with lower training loss, lower peak memory, and higher prefill and decoding throughput.
        (c) Final architecture hyperparameters of MNN-Pocket-0.3B.
    }
    \label{fig:architecture_search}
\end{figure}

We select the model architecture through a two-stage process. We first determine the overall parameter scale based on deployment memory and inference speed, and then, within the fixed parameter budget, jointly optimize the internal configuration for model capability and hardware efficiency.
We scale the dense model architecture from $\sim$50M to $\sim$1.3B parameters and deploy each candidate on an M5-based iPad Pro using the MNN Metal backend~\cite{jiang2020mnn} under identical inference settings.
As shown in Figure~\ref{fig:architecture_search}(a), larger models incur higher peak memory usage and lower decoding throughput. We set a 300\,MB peak-memory budget to limit the inference footprint under mobile hardware and application constraints, together with a 200 tokens/s decoding target for real-time interaction. Under W4A16 quantization and representative input and output lengths of 256 tokens, the approximately 300M-parameter model is the largest candidate satisfying both constraints. We therefore select this scale and optimize its internal architecture in the second stage.

Within the fixed 300M-parameter budget, we further optimize the internal architecture by jointly varying depth and width while adjusting the feed-forward and attention dimensions to maintain comparable model sizes.
Figure~\ref{fig:architecture_search}(b) compares representative 16-, 28-, and 36-layer configurations. Increasing depth from 16 to 28 layers substantially reduces training loss and improves evaluation performance, with only moderate memory and decoding overhead. Further increasing depth to 36 layers yields no additional capability gain, while reducing decoding throughput from 355 to 323 tokens/s. We therefore select the 28-layer configuration, which approaches the capability of the deeper model while providing a more favorable efficiency trade-off.
The final configuration of MNN-Pocket-0.3B is summarized in Figure~\ref{fig:architecture_search}(c).

\subsection{Tokenizer Design}

Tokenizer design is particularly consequential for on-device LLMs. A larger vocabulary improves sequence compression but increases the parameter and memory costs of the embedding and output layers, which constitute a substantial fraction of a sub-billion-parameter model. It also enlarges the output projection and therefore increases decoding latency. To balance tokenization efficiency with deployment cost, we derive a compact 64K vocabulary from the Qwen3 tokenizer~\cite{Qwen3} while preserving its coverage of Chinese, English, and specialized domains.


We construct a 6-GiB corpus spanning pre-training, supervised fine-tuning, and constraint-following data, with broad coverage of Chinese, English, code, mathematics, and instruction-oriented formats. After tokenizing the corpus with the original tokenizer, we rank tokens by frequency and greedily select high-frequency entries under the vocabulary budget. All special tokens and 256 byte-fallback tokens are retained, together with the complete BPE merge-dependency chain of every selected token. This dependency-aware pruning preserves valid segmentation and closely approximates the behavior of the original tokenizer without introducing the instability or initialization mismatch associated with training a new tokenizer from scratch.

Finally, we remap the retained tokens to a contiguous ID space. The resulting tokenizer reduces the vocabulary from approximately 151K to 64K tokens—a reduction of about 57\%—thereby substantially decreasing embedding and output-layer parameters as well as vocabulary-dependent decoding overhead. Meanwhile, it retains effective tokenization across Chinese and English text, code, mathematics, and instruction-following data, providing a favorable capability–efficiency trade-off for on-device deployment.

\begin{figure}[!t]
    \centering
    \includegraphics[width=\linewidth]{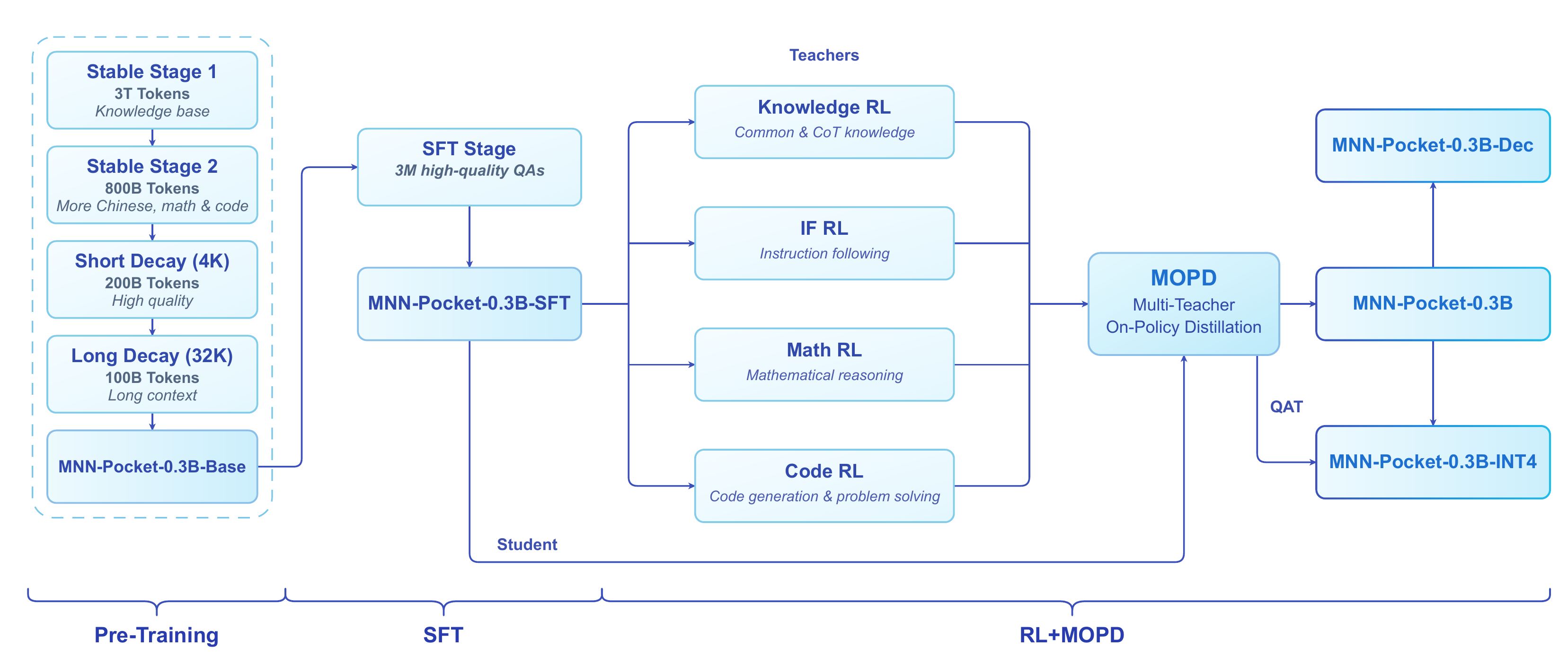}
    \caption{\textbf{Training pipeline overview.} Four-stage pre-training produces MNN-Pocket-0.3B-Base, followed by supervised fine-tuning (SFT) on 3 million high-quality question--answer pairs to obtain MNN-Pocket-0.3B-SFT. Four domain-specialized reinforcement learning (RL) teachers covering knowledge, instruction following, mathematics, and code guide MOPD of the SFT student to obtain MNN-Pocket-0.3B. Quantization-aware training (QAT) further converts the consolidated model into MNN-Pocket-0.3B-INT4 for on-device deployment. The consolidated model also initializes MNN-Pocket-0.3B-Dec for non-autoregressive decision modeling.}
    \label{fig:method_overview}
\end{figure}

Figure~\ref{fig:method_overview} summarizes the training pipeline built on the architecture and tokenizer described above. The following sections detail the pre-training and post-training stages.





\section{Pre-Training}

In the sub-billion-parameter regime, better data engineering and training schedules matter much more than simply increasing training tokens. This section outlines the principles guiding our pre-training design, then presents the data engineering pipeline, and finally describes the training recipe for MNN-Pocket-0.3B-Base.

\subsection{Principles for On-device LLM Pre-training}

Tight parameter budgets impose severe capacity constraints, making model capability more sensitive to data quality, difficulty, and domain balance. Our objective is therefore to curate, synthesize, and organize higher-quality data under a fixed tokens-per-parameter (TPP) ratio. The pre-training design of MNN-Pocket-0.3B-Base is guided by the following principles:
\begin{itemize}
    \item \textbf{Quality over Scale.} 
    On-device LLMs are less tolerant to noisy, redundant, or weakly informative tokens, making data quality rather than scale a primary determinant of final capability. 
    \item \textbf{Controlled Knowledge Density and Difficulty.} 
    Effective pre-training data should be not only high-quality but also sufficiently informative and appropriately learnable. We therefore control data attributes such as knowledge density and difficulty to support more precise filtering, analysis, construction, and mixture.
    \item \textbf{Stage-aware Domain Balance.} Since domains compete more strongly for representation in capacity-limited small LLMs, and different domains often favor different training stages and settings, effective pre-training especially requires finer-grained control to achieve a well-balanced domain composition, with mixtures adjusted across training stages.
\end{itemize}



\begin{figure}[!t]
    \centering
    \includegraphics[width=\textwidth]{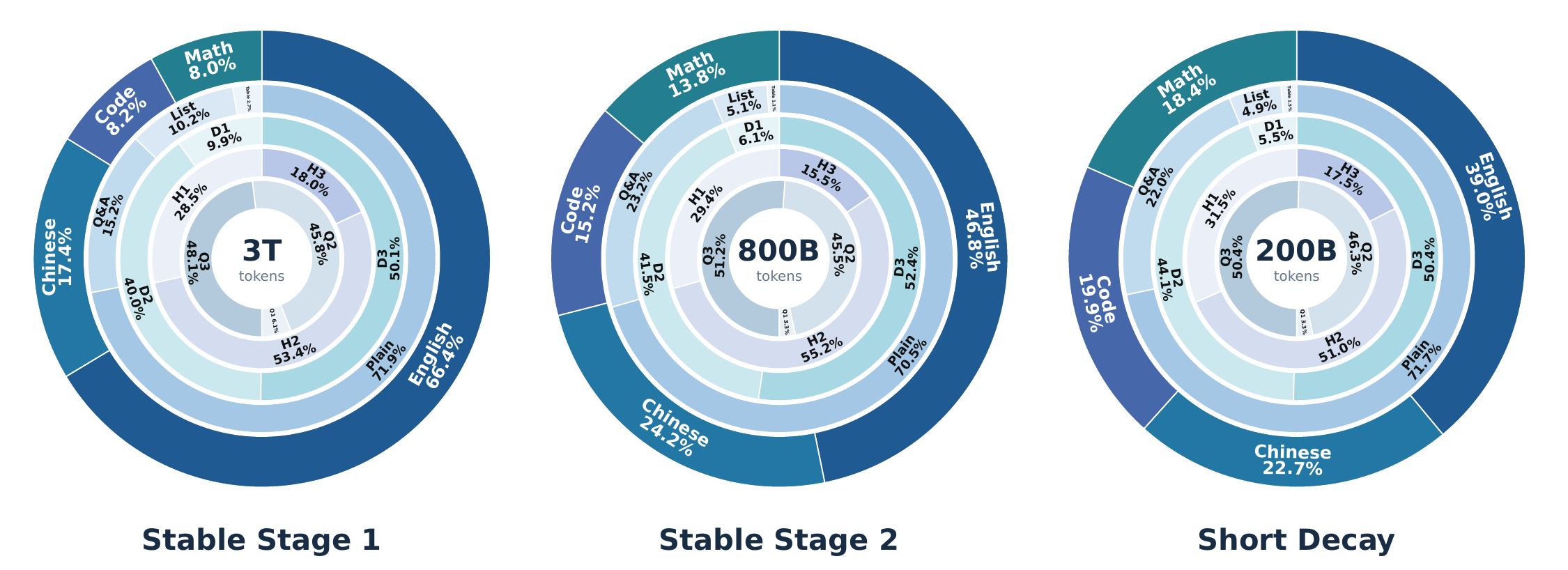}
    \caption{
        \textbf{Pre-training data composition across training stages.}
        From outer to inner, the concentric rings show domain, style, knowledge density, difficulty, and quality distributions. Knowledge density (D1–D3), difficulty (H1–H3), and quality (Q1–Q3) are each grouped into three levels, from low to high. Percentages are normalized independently within each dimension using recorded token weights. Center labels indicate the training token budget for each stage, in trillions (T) or billions (B).
    }
    \label{fig:stagewise_data_distribution}
\end{figure}

\subsection{Multi-dimensional Pre-training Data Engineering}

Following these principles, we build a pre-training pipeline that treats data as a structured, schedulable asset rather than a uniformly filtered corpus. After cleaning and deduplication, we apply structure-aware rewriting and normalization to semi-structured sources, including question-answer pairs, tables, and lists, to expose their underlying semantics. For Math and Code, we use frontier models to enrich selected samples with relevant concepts, intermediate reasoning steps, and explanatory comments. This processing improves knowledge density and learnability, allowing a capacity-limited model to acquire useful structure from fewer tokens.

To support fine-grained control, each document is annotated with the following attributes:
\begin{itemize}
    \item \textbf{Domain} classifies each document hierarchically, covering broad domains such as English, Chinese, Math, and Code, and finer-grained subdomains (e.g., algebra and geometry within Math).
    \item \textbf{Quality} reflects the general cleanliness, coherence, and expected learning value of the sample.
    \item \textbf{Knowledge density} uses signals such as content diversity, entity and concept density, and redundancy to distinguish informative content from verbose or weakly informative data.
    \item \textbf{Difficulty} measures sample complexity using signals such as prerequisite level, reasoning depth, and domain-specific complexity.
    \item \textbf{Style} captures format-related properties such as plain prose, QA, structured lists, or tabularized text.
\end{itemize}

Rather than collapsing these properties into a single scalar score, we use them to form multi-dimensional joint buckets over quality, density, difficulty, style, and domain. This allows the corpus to be analyzed and mixed at a much finer granularity than conventional source-level or score-threshold filtering. During pre-training, these buckets become the basic units for stage-wise mixture design, enabling us to determine not only whether a sample is useful, but also when it is useful and in what proportion.

\subsection{Pre-training Recipe}

Pre-training a 300M-parameter model requires tighter coordination between data composition and optimization than simply scaling down recipes developed for larger LLMs. In particular, domain proportions cannot be optimized independently of their internal quality, knowledge-density, and difficulty distributions: changing these attributes within one domain can alter its optimal share relative to other domains. We therefore formulate pre-training recipe design as a joint search over the complete multi-dimensional data distribution, together with stage boundaries, learning rates, and token budgets.


For each stage \(t\), we partition the corpus into joint buckets \(b=(m,q,d,h,s)\), corresponding to domain, quality, knowledge density, difficulty, and style, and define its data recipe as a sampling distribution \(p_t(b)\). Rather than tuning individual dimensions sequentially, we jointly search the bucket weights, learning-rate schedule, token budget, and stage boundary.
Let \(\theta_t^{(0)}\) denote the checkpoint entering stage \(t\). From this shared initialization, short proxy runs train models on different candidate recipes under controlled budgets, yielding proxy checkpoints
\(\theta_t^{(i)}\) and updates
\(\Delta\theta_t^{(i)}=\theta_t^{(i)}-\theta_t^{(0)}\). We interpolate these
updates as
\begin{equation}
\widetilde{\theta}_t(\boldsymbol{\alpha})
=\theta_t^{(0)}+\sum_i\alpha_i\Delta\theta_t^{(i)},
\qquad \sum_i\alpha_i=1,
\end{equation}
to coarsely screen a broader mixture space. Since model interpolation only
approximates the trajectory of training on an actual mixture, it is not used
for final recipe selection. Instead, we perform a fine-grained search around
the strongest candidates by training on their actual bucket-level mixtures
and jointly refining the sampling weights and optimization settings. The final candidates are evaluated using convergence, validation loss, downstream capability, and cross-domain regression.


This procedure yields a highly data-efficient pre-training recipe that achieves frontier-level performance at the 300M scale, as demonstrated in Section~\ref{sec:evaluation}.
Figure~\ref{fig:stagewise_data_distribution} and Table~\ref{tab:mobilemoe-hparams} summarize the multi-dimensional data composition and optimization hyperparameters, respectively, across the pre-training stages of MNN-Pocket-0.3B-Base.

\begin{table}[!t]
\centering
\caption{Training hyperparameters for each pre-training stage of MNN-Pocket-0.3B-Base.}
\label{tab:mobilemoe-hparams}

\begin{threeparttable}
\renewcommand{\arraystretch}{1.15}
\setlength{\tabcolsep}{10pt}

\begin{tabular}{lcccc}
\toprule
\textbf{Hyperparameter} & \textbf{Stable Stage 1} & \textbf{Stable Stage 2} & \textbf{Short Decay} & \textbf{Long Decay} \\
\midrule
Context length & 4,096 & 4,096 & 4,096 & 32,768 \\
Total tokens & $\sim$3T & $\sim$800B & $\sim$200B & $\sim$100B \\
Global batch size & 1024 & 1024 & 1024 & 128 \\
Tokens per step & 4.2M & 4.2M & 4.2M & 4.2M \\
Peak learning rate & $3 \times 10^{-4}$ & $1 \times 10^{-4}$ & $4 \times 10^{-5}$ & $4 \times 10^{-5}$ \\
LR schedule & Constant & Constant & Cosine & Cosine \\
Warmup steps & 3,000 & 3,000 & 1000 & 1000 \\
Hardware (GPUs) & 32 & 32 & 32 & 32 \\
Wall-clock time & 7--8 weeks & $\sim$2 weeks & \textasciitilde2--3 days & \textasciitilde2--3 days \\
\midrule
Optimizer & \multicolumn{4}{c}{AdamW ($\beta_1 = 0.9$, $\beta_2 = 0.95$, $\epsilon = 10^{-15}$)} \\
Weight decay & \multicolumn{4}{c}{0.1} \\
Gradient clipping & \multicolumn{4}{c}{1.0} \\
\bottomrule
\end{tabular}

\end{threeparttable}
\end{table}


\section{Post-Training}
\label{sec:posttraining}

We adapt MNN-Pocket-0.3B-Base for on-device instruction following through a three-stage post-training pipeline: supervised fine-tuning with the SFT Loop Engine, multi-teacher on-policy distillation for capability consolidation, and quantization-aware training for efficient low-precision deployment.

\subsection{Supervised Fine-Tuning (SFT)}
\label{sec:foundational-sft-data}

Foundational SFT combines curated open-source data with targeted synthesis from the SFT Loop Engine (Section~\ref{sec:sft-loop-engine}) to form the final training mixture, supporting on-device instruction following while aligning with foundational capabilities and deployment scope.


\begin{figure}[!b]
    \centering
    \includegraphics[width=\linewidth]{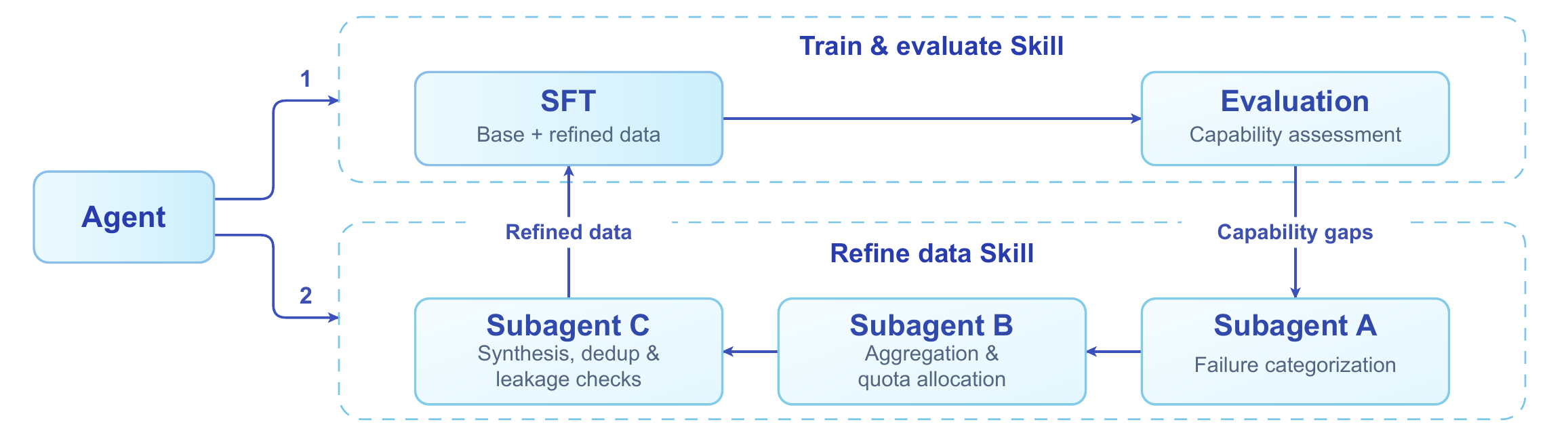}
    \caption{\textbf{SFT Loop Engine.} An agent invokes the training/evaluation (1) and data refinement (2) skills. Subagents categorize failures (A), aggregate results and allocate quotas (B), and synthesize, deduplicate, and screen data for leakage (C). The refined data feeds the next SFT iteration.}
    \label{fig:sft-loop-engine-pipeline}
\end{figure}

\iflatexml
  \newcommand{\sftHeadingRowColor}{\rowcolor[HTML]{EAF1FC}}
\else
  \newcommand{\sftHeadingRowColor}{\rowcolor[HTML]{EAF1FC}[8pt][8pt]}
\fi

\begin{table}[!t]
\centering
\caption{Token-based composition of the final filtered SFT data recipe across taxonomy families and capabilities.}
\label{tab:sft-data-composition}
\label{tab:sft-family-composition}
\label{tab:sft-capability-composition}
\small
\setlength{\tabcolsep}{4pt}
\renewcommand{\arraystretch}{1.04}
\begin{tabular}{@{\hspace{8pt}}lrrr@{\hspace{8pt}}}
\toprule
\textbf{Category} & \textbf{Samples} & \textbf{Tokens} & \textbf{Percent} \\
\midrule
\sftHeadingRowColor
\multicolumn{4}{@{\hspace{8pt}}l@{\hspace{8pt}}}{\textbf{Family}} \\
\addlinespace[1pt]
Multilingual Creation \& Open-Ended Generation & 607K & 289M & $15.81\%$ \\
Mathematics \& Quantitative Reasoning & 722K & 393M & $21.54\%$ \\
Programming \& Algorithmic Coding & 801K & 433M & $23.70\%$ \\
Application Development \& General QA & 247K & 148M & $8.08\%$ \\
General Chinese-Language Tasks & 450K & 337M & $18.43\%$ \\
STEM QA \& Scientific Computation & 334K & 227M & $12.45\%$ \\
\addlinespace[2pt]
\cmidrule(lr){1-4}
\addlinespace[2pt]
\sftHeadingRowColor
\multicolumn{4}{@{\hspace{8pt}}l@{\hspace{8pt}}}{\textbf{Capability}} \\
\addlinespace[1pt]
Code & 970K & 558M & $30.58\%$ \\
Knowledge QA & 747K & 482M & $26.41\%$ \\
Mathematics & 799K & 443M & $24.26\%$ \\
Creative Writing & 378K & 250M & $13.67\%$ \\
Instruction Following & 90.0K & 23.6M & $1.29\%$ \\
Others & 178K & 69.2M & $3.79\%$ \\
\midrule
\textbf{Total} & \textbf{3.16M} & \textbf{1.83B} & $\mathbf{100.00\%}$ \\
\bottomrule
\end{tabular}
\end{table}

\subsubsection{SFT Data Mixture}
\label{sec:sft-data-mixture}

The core corpus is drawn from open-source datasets covering mathematical reasoning, programming, knowledge QA, and open-ended generation. We supplement it with 116.6K targeted synthetic examples from the SFT Loop Engine. After data filtering, the resulting mixture provides broad task coverage while supporting targeted additions for capability gaps. The final recipe contains approximately 3.16M samples and 1.83B assistant tokens, with all percentages normalized by assistant token count. Table~\ref{tab:sft-data-composition} summarizes the mixture from two complementary perspectives: taxonomy family and capability.

Our post-training data system uses a three-level taxonomy comprising 1{,}511 fine-grained categories, 154 mid-level categories, and six families, providing a consistent basis for data versioning, quota assignment, and synthesis.

\subsubsection{SFT Loop Engine}
\label{sec:sft-loop-engine}

The compact size of MNN-Pocket-0.3B keeps the resource cost of supervised fine-tuning (SFT) modest, making repeated training and refinement practical. Building on this, we develop the SFT Loop Engine to explore recursive self-improvement (RSI) through iterative cycles of training, evaluation, and targeted data synthesis.

As shown in Figure~\ref{fig:sft-loop-engine-pipeline}, a lightweight main agent coordinates two skills: Train \& evaluate and Refine data. The training and evaluation skill assembles the current data mixture from base and incremental or replacement data, trains an SFT checkpoint, and collects evaluation results. Within the data refinement skill, Subagent~A uses the post-training data system to map failed responses and their generation traces to fine-grained categories. Subagent~B aggregates failure attributions and historical training-data statistics to assign per-category and per-language synthesis quotas. Subagent~C generates targeted synthetic examples without access to the original evaluation items, then performs deduplication and leakage screening. The main agent maintains only workflow state and model-version decisions, and the resulting data update is returned to the training skill to close the loop for the next SFT iteration.

\FloatBarrier

\subsection{Multi-Teacher On-Policy Distillation}

Building on the SFT student, the remaining challenge is to strengthen
code, mathematical reasoning, knowledge, and instruction following
within a fixed on-device parameter budget. Joint multi-domain reinforcement learning (RL)
must reconcile potentially competing updates, while sequential
domain-wise tuning risks making the final policy dependent on training
order. We therefore adopt Multi-Teacher On-Policy Distillation (MOPD) to consolidate domain experts independently trained through RL into a single student on student-generated trajectories, as shown in Step~1 of Figure~\ref{fig:mopd_framework}. The consolidated BF16 student is then compressed to a low-precision student and recovered by Self-OPD with quantization-aware training, shown in Step~2 of the same figure.

\subsubsection{Domain-Specialized Teachers and Routing}
\label{sec:mopd-teachers}

We construct four domain-specialized teacher routes: code,
mathematics, knowledge, and instruction following, listed as the four
same-origin teachers inside the routed-teacher box of Step~1 in
Figure~\ref{fig:mopd_framework}. The teachers are
initialized from a common base model and further optimized using
domain-specific RL objectives. They share the same architecture,
tokenizer, vocabulary, and chat template, which reduces distribution
mismatch during distillation.

Each training sample inherits a domain label $d$ from its source
dataset, which deterministically selects one frozen teacher $q_d^T$.
Within a mixed-domain batch, samples are grouped by label for teacher
inference, and the returned logits are restored to their original
order to supervise the shared student.

\subsubsection{GKD-Style OPD with Reverse KL}
\label{sec:mopd_gkd}
OPD typically uses either a PG objective or a GKD objective. The PG
formulation uses the difference between the teacher and student log
probabilities of the sampled token as a detached advantage. GKD instead
matches their token distributions at each prefix generated by the student,
providing supervision over multiple candidate tokens. We use GKD because
this denser signal has lower variance and provides more stable training
and better performance for our compact student.

For the distillation loss, we choose reverse KL for distribution matching. Unlike the mode-covering
forward KL, reverse KL emphasizes the teacher's dominant modes, enabling
more selective knowledge transfer to the capacity-constrained student.
Let $\mathcal{V}_{d,t}^{K}$ denote the teacher's Top-$K$ candidates, and
let $\bar{p}_{\theta,t}$ and $\bar{q}_{d,t}^{T}$ be the student and teacher
distributions renormalized over this set. The token-level distillation
loss is
\begin{equation}
\ell_t^{\mathrm{RKL}}
=
D_{\mathrm{KL}}
\left(
\bar{p}_{\theta,t}
\,\Vert\,
\bar{q}_{d,t}^{T}
\right)
=
\sum_{v \in \mathcal{V}_{d,t}^{K}}
\bar{p}_{\theta,t}(v)
\log
\frac{\bar{p}_{\theta,t}(v)}
     {\bar{q}_{d,t}^{T}(v)}.
\end{equation}

The overall MOPD objective is
\begin{equation}
\mathcal{L}_{\mathrm{MOPD}}(\theta)
=
\mathbb{E}_{\substack{(x,d)\sim\mathcal{D}\\
y\sim p_\theta(\cdot\mid x)}}
\left[
\frac{1}{|y|}
\sum_{t=1}^{|y|}
\ell_t^{\mathrm{RKL}}
\right].
\end{equation}

\subsubsection{Self-OPD with Quantization-Aware Training}
\label{sec:quantization-aware-training}

\begin{figure}[!t]
    \centering
    \includegraphics[width=\textwidth]{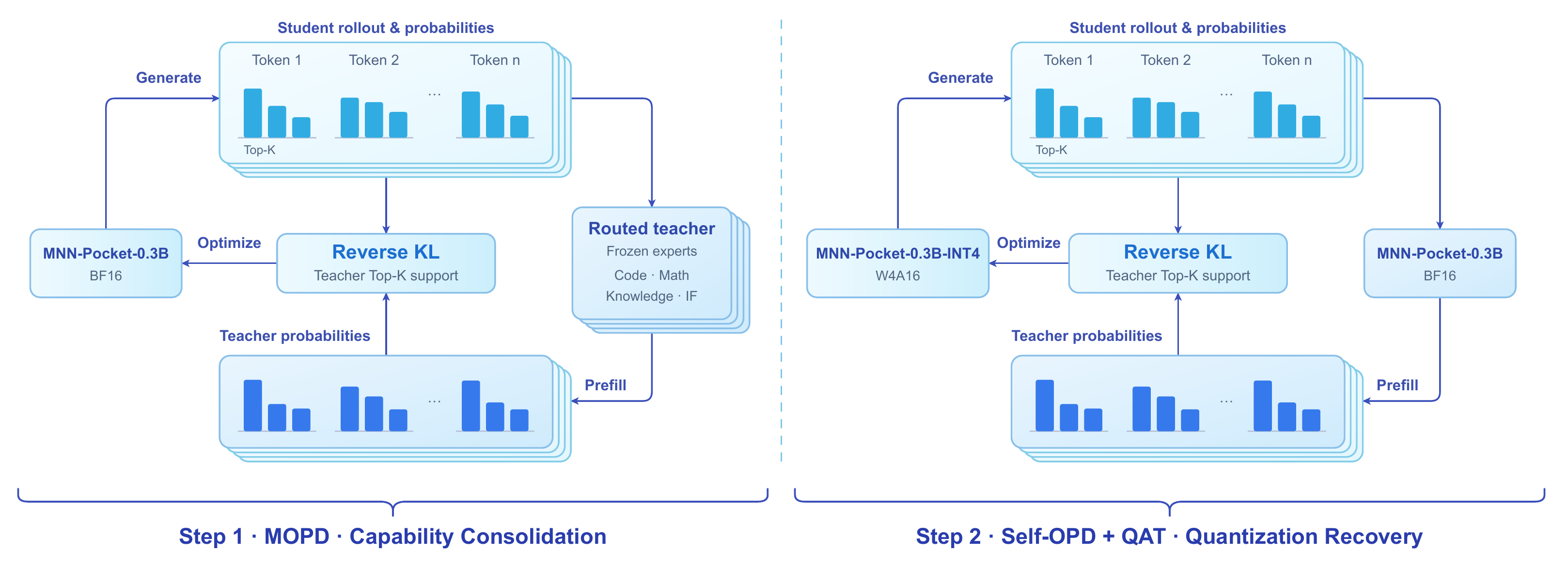}
    \caption{
        \textbf{Two-step consolidation and low-precision recovery.}
        Both steps run the same on-policy distillation loop.
        \textbf{Step~1} (left): Multi-Teacher On-Policy Distillation (MOPD),
        in which four same-origin RL teachers---code, mathematics, knowledge,
        and instruction following---are routed by domain label and the selected
        teacher prefills the MNN-Pocket-0.3B student's own rollout, so that
        domain-routed Top-$K$ reverse-KL distillation supervises the student in
        BF16.
        \textbf{Step~2} (right): Self-OPD with quantization-aware training
        (QAT), which instead uses a single frozen teacher: MNN-Pocket-0.3B in
        BF16 serves as the self-teacher, and MNN-Pocket-0.3B-INT4 in W4A16 form,
        initialized by post-training quantization, is aligned to it on its own
        rollout.
    }
    \label{fig:mopd_framework}
\end{figure}

After MOPD, we freeze the consolidated BF16 model as a self-teacher and initialize a low-precision student from the same checkpoint through post-training quantization (PTQ), as shown in Step~2 of Figure~\ref{fig:mopd_framework}. The student uses MNN-compatible asymmetric INT4 weight quantization with a group size of 64 and 16-bit activations (W4A16), with shared quantized weights for the token embedding and LM head. To mitigate quantization-induced degradation, we combine Self-OPD with quantization-aware training (QAT), aligning the student and frozen teacher distributions on student-generated prefixes using the Top-$K$ reverse-KL objective in Section~\ref{sec:mopd_gkd}. Only the student is updated and retained for deployment.

\subsection{Non-Autoregressive Decision Modeling}
\label{sec:non-autoregressive-decision-modeling}

Following Kev and Laya~\citep{kev_decision_model,laya_decision_model}, we adapt
MNN-Pocket-0.3B into \textbf{MNN-Pocket-0.3B-Dec}, a decision model using a shared
causal Transformer backbone, task-specific LoRA adapters, and a pointer-based decision head.

\begingroup
\setlength{\intextsep}{6pt}
\begin{figure}[H]
    \centering
    \includegraphics[width=\linewidth]{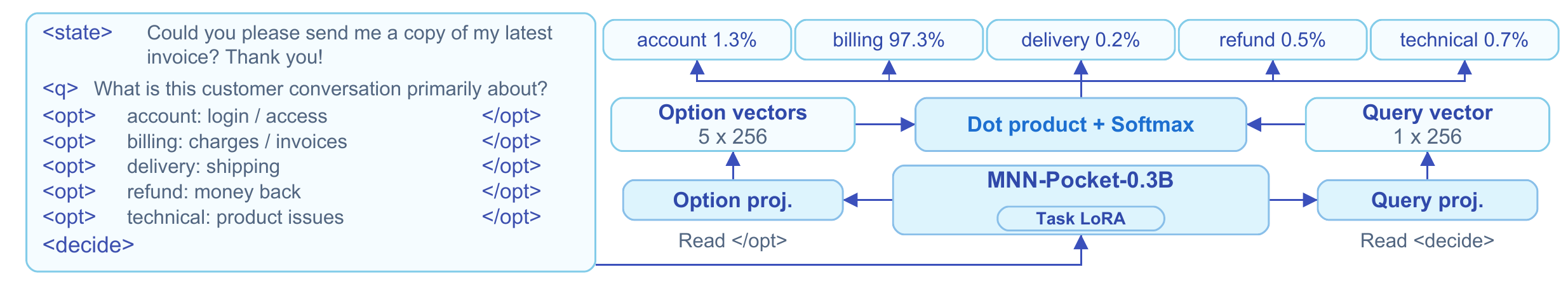}
    \caption{\textbf{MNN-Pocket-0.3B-Dec architecture.} Left: abridged
    Typed Decisions~\citep{typed_decisions_dataset} customer-service input.
    Right: the shared model architecture with dataset reference probabilities
    shown for illustration. Markers are semantic aliases of reserved tokens.}
    \label{fig:non-autoregressive-decision-architecture}
\end{figure}
\endgroup

Let $\mathbf h_{\mathrm{dec}}$ and $\mathbf h_i$ be final-RMSNorm states at the
decision and option-readout tokens. Shared affine projections and scaled
dot-product normalization give
\begin{equation}
\label{eq:decision-head-probabilities}
\mathbf q=W_Q\mathbf h_{\mathrm{dec}}+\mathbf b_Q,\quad
\mathbf k_i=W_K\mathbf h_i+\mathbf b_K,\quad
p_i=\operatorname{softmax}_i\!\left(
\left[\mathbf q^\top\mathbf k_j/\sqrt{d_p}\right]_{j=1}^{M}\right).
\end{equation}
Here $W_Q,W_K$ and $\mathbf b_Q,\mathbf b_K$ are projection weights and biases;
$d_p=256$ is the projection width, $M$ is the valid-option count, and $p_i$ is option $i$'s probability.
The decision path reads all outputs in one forward pass, bypassing the retained
LM head and token-by-token decoding.

\iflatexml\else
  \Needspace{6\baselineskip}
\fi
\section{On-Device Inference and Deployment}

\subsection{Efficient Inference with MNN}

MNN-Pocket-0.3B is deployed on MNN \cite{jiang2020mnn,mnn-llm}, an open-source on-device inference engine maintained by the MNN team. At runtime, an input prompt is first encoded by the model tokenizer and processed in a prompt-prefill pass that materializes the KV cache. Generation then proceeds autoregressively, consuming one token at a time while reusing the cached states. The generated token IDs are incrementally decoded and streamed to the application. This execution path
requires no network connection and keeps both model computation and user data on the device.

The two inference phases have different computational characteristics. Prefill exposes parallelism across the
prompt and is dominated by matrix--matrix multiplication and attention, whereas batch-one decoding repeatedly
streams the model weights and is more sensitive to memory bandwidth and kernel-dispatch overhead. MNN therefore
uses phase-aware low-bit projection kernels, fused Transformer operators, and backend-specific attention and
KV-cache implementations. Beyond these kernel-level optimizations, MNN also supports speculative decoding, which we apply to on-device generation in Section~\ref{sec:spec}. The exported MNN graph provides a common integration interface across devices, while
the runtime maps its operators to the available hardware backend. Dedicated NPU deployment may additionally
require backend-specific graph conversion, calibration, and compiled model assets.

Unless otherwise stated, the deployment results in this section use MNN-Pocket-0.3B-INT4, obtained through the quantization-aware training procedure in Section~\ref{sec:quantization-aware-training}. Here we focus on the systems impact of quantization---model size,
runtime memory, and inference latency.

\subsection{On-Device Efficiency Evaluation}

We evaluate on-device language models on the iPad Pro M5 and Snapdragon 8 Gen 5 using the MNN Metal, OpenCL, and CPU backends. The models run on MNN 3.6.2 with four CPU threads, low-precision and low-memory modes, and INT4 quantization. Greedy decoding is used to reduce sampling randomness and improve reproducibility. Following llama-bench, each configuration is warmed up once and then measured five times. The tables report the median throughput; prefill and decode are measured independently from an empty KV cache.

\textbf{Memory and Deployment Footprint}\quad
Table~\ref{tab:m5-metal-file-size-peak-memory-context512} compares the deployment footprint and peak Metal resident memory usage of each model at a context length of 512 tokens. The MNN deployment footprint includes model weights, the tokenizer, the computation graph, and inference configuration. Peak memory is measured as the maximum Metal resident memory during model loading and inference, minus the baseline measured before model loading. On the iPad Pro (M5), MNN-Pocket-0.3B requires a deployment footprint of 173 MB and reaches a peak memory usage of 228 MB above baseline.
 
\textbf{Inference Throughput}\quad
Table~\ref{tab:on-device-pp512-tg512} reports inference throughput for 512 token prefill and decode on the iPad Pro M5 and Snapdragon 8 Gen 5. On the iPad Pro M5, MNN-Pocket-0.3B achieves 8,721 tok/s prefill and 466 tok/s decode with Metal, and 2,228 tok/s prefill and 231 tok/s decode with CPU. On Snapdragon 8 Gen 5, it achieves 3,403 tok/s prefill and 138 tok/s decode with OpenCL, and 930 tok/s prefill and 251 tok/s decode with CPU.

Overall, MNN-Pocket-0.3B combines a small deployment footprint and low peak memory usage with high prefill and decode throughput, demonstrating strong efficiency for on-device deployment.

\begin{table}[!b]
  \centering
  \footnotesize
  \setlength{\tabcolsep}{3pt}
  \renewcommand{\arraystretch}{1.15}
  \caption{Memory and deployment footprint of on-device LLMs on iPad Pro (M5) at 512 token context.}
  \label{tab:m5-metal-file-size-peak-memory-context512}
  \resizebox{0.95\textwidth}{!}{%
  \begin{tabular}{lcccccccc}
    \toprule
    \raisebox{0.5\baselineskip}{\strut Metric}
      & \shortstack{MNN-Pocket\\0.3B}
      & \shortstack{Gemma-3\\270M}
      & \shortstack{ERNIE-4.5\\0.3B}
      & \shortstack{LFM2.5\\350M}
      & \shortstack{SmolLM2\\360M}
      & \shortstack{MiniCPM4\\0.5B}
      & \shortstack{Hunyuan\\0.5B}
      & \shortstack{Qwen3\\0.6B} \\
    \midrule
    MNN inference file size (MB)
      & 173 & 162 & 207 & 202 & 206 & 247 & 307 & 340 \\
    Peak Memory (MB)
      & 228 & 272 & 287 & 258 & 257 & 305 & 449 & 427 \\
    \bottomrule
  \end{tabular}%
  }
\end{table}

\begin{table}[!t]
  \centering
  \footnotesize
  \setlength{\tabcolsep}{3pt}
  \renewcommand{\arraystretch}{1.15}
  \caption{Inference speed (tok/s) of on-device LLMs across different platforms at 512 token context.}
  \label{tab:on-device-pp512-tg512}
  \resizebox{\textwidth}{!}{%
  \begin{tabular}{lllcccccccc}
    \toprule
    \raisebox{0.5\baselineskip}{\strut Device} & \raisebox{0.5\baselineskip}{\strut Backend} & \raisebox{0.5\baselineskip}{\strut Stage}
      & \shortstack{MNN-Pocket\\0.3B}
      & \shortstack{Hunyuan\\0.5B}
      & \shortstack{Qwen3\\0.6B}
      & \shortstack{MiniCPM4\\0.5B}
      & \shortstack{SmolLM2\\360M}
      & \shortstack{ERNIE-4.5\\0.3B}
      & \shortstack{LFM2.5\\350M}
      & \shortstack{Gemma-3\\270M} \\
    \midrule
    \multirow{4}{*}{iPad Pro (M5)}
      & \multirow{2}{*}{Metal} & Prefill
      & 8721 & 5330 & 6027 & 6973 & 7067 & 7218 & 8026 & 9519 \\
      & & Decode
      & 466 & 262 & 286 & 357 & 397 & 395 & 504 & 455 \\
    \cmidrule(lr){2-11}
      & \multirow{2}{*}{CPU} & Prefill
      & 2228 & 1474 & 1396 & 1818 & 1824 & 2185 & 2234 & 3793 \\
      & & Decode
      & 231 & 149 & 135 & 197 & 187 & 224 & 262 & 290 \\
    \midrule
    \multirow{4}{*}{Android (8 Gen 5)}
      & \multirow{2}{*}{OpenCL} & Prefill
      & 3403 & 1757 & 1627 & 2368 & 2291 & 2532 & 3070 & 4209 \\
      & & Decode
      & 138 & 88 & 93 & 117 & 127 & 119 & 175 & 118 \\
    \cmidrule(lr){2-11}
      & \multirow{2}{*}{CPU} & Prefill
      & 930 & 556 & 522 & 682 & 719 & 853 & 889 & 1953 \\
      & & Decode
      & 251 & 137 & 126 & 189 & 199 & 216 & 249 & 293 \\
    \bottomrule
  \end{tabular}%
  }
\end{table}
\subsection{Speculative Decoding}
\label{sec:spec}

To further improve batch-one decoding efficiency, we combine weight
quantization with DFlash2~\citep{dflash2}. The draft proposes a block of
tokens in parallel, which the target then verifies in a single forward
pass~\citep{speculative}.

\textbf{Draft architecture selection}\quad
Methods such as EAGLE~\citep{EAGLE}, DFlash~\citep{dflash}, and
DSpark~\citep{dspark} use target hidden states to construct proposals and
apply a language-model head to obtain token distributions. With a compact
0.3B target, minimizing draft computation and vocabulary-projection
overhead is particularly important. An autoregressive draft requires a
draft forward pass and an LM-head computation for each proposed token.
DFlash2 instead generates a proposal block in one draft forward pass and
computes logits for all proposal positions in a batch. Our implementation
shares the target's token embedding and LM head and uses the
position-weighted loss of D-PACE~\citep{dpace}.

\textbf{Draft training}\quad
We train DFlash2 drafts using either the full-precision MNN-Pocket-0.3B
model or its QAT counterpart to provide training signals,
yielding the MNN-trained draft and the MNN-INT4-trained draft,
respectively. Training is built on SpecForge~\citep{specforge}.
Both pipelines use prompts from Open-PerfectBlend~\citep{perfectblend}
and 500K post-training samples with the corresponding target. Unless otherwise stated,
speculative-decoding experiments use block size $B{=}8$, batch size 1,
and greedy decoding.
We report mean accepted length per decoding round, denoted AL, and
on-device decode throughput. Speedup is measured against autoregressive
decoding with the same target, backend, and workload. We use six
prompt-only benchmarks: MATH-500~\citep{math500} and
AIME25~\citep{aime25} for math, MBPP~\citep{mbpp} and
HumanEval~\citep{humaneval} for code, and MT-Bench~\citep{mtbench}
and Arena-Hard~\citep{arenahard} for chat.

\begin{table}[!b]
\centering
\footnotesize
\setlength{\tabcolsep}{3.0pt}
\renewcommand{\arraystretch}{1.05}
\caption{Mean accepted length across draft–target configurations for MNN-Pocket-0.3B.}
\label{tab:dflash2}
\resizebox{\linewidth}{!}{%
\begin{tabular}{lllccccccc}
\toprule
\multirow[b]{2}{*}{\shortstack[l]{Draft\\Precision}}
& \multirow[b]{2}{*}{\shortstack[l]{Training\\Source}}
& \multirow[b]{2}{*}{\shortstack[l]{Verification\\Model}}
& \multicolumn{2}{c}{Math}
& \multicolumn{2}{c}{Code}
& \multicolumn{2}{c}{Chat}
& \multirow{2}{*}{Avg.} \\
\cmidrule(lr){4-5}\cmidrule(lr){6-7}\cmidrule(lr){8-9}
& & & MATH-500 & AIME25 & MBPP & HumanEval & MT-Bench & A-Hard & \\
\midrule
BF16 & MNN & MNN
& 5.34 & 4.59 & 5.55 & 5.83 & 3.55 & 3.18 & 4.67 \\
BF16 & MNN & MNN-INT4
& 5.30 & 4.49 & 5.57 & 5.75 & 3.62 & 3.16 & 4.65 \\
BF16 & MNN-INT4 & MNN-INT4
& 5.34 & 4.55 & 5.64 & 5.82 & 3.70 & 3.18 & 4.70 \\
INT4 & MNN & MNN-INT4
& 5.16 & 4.37 & 5.48 & 5.65 & 3.50 & 3.06 & 4.54 \\
INT4 & MNN-INT4 & MNN-INT4
& 5.22 & 4.47 & 5.55 & 5.76 & 3.65 & 3.10 & 4.63 \\
\bottomrule
\end{tabular}%
}
\end{table}

\begin{table}[!t]
\centering
\footnotesize
\setlength{\tabcolsep}{4pt}
\renewcommand{\arraystretch}{1.05}
\caption{Performance with MNN-Pocket-0.3B MNN-INT4 target and the INT4 MNN-trained draft.}
\label{tab:dflash2-device}
\begin{tabular}{llcccc}
\toprule
Device & Workload & AR tok/s & Mean AL & Spec. tok/s & Speedup \\
\midrule
\multirow{4}{*}{iPad Pro (M5)}
& Math & \multirow{4}{*}{435} & 4.49 & 709 & 1.63$\times$ \\
& Code & & 5.73 & 895 & 2.06$\times$ \\
& Chat & & 3.51 & 547 & 1.26$\times$ \\
& Avg. & & 4.57 & 717 & 1.65$\times$ \\
\midrule
\multirow{4}{*}{Android (8 Gen 5)}
& Math & \multirow{4}{*}{135}& 4.48 & 159& 1.17$\times$\\
& Code & & 5.72 & 186& 1.38$\times$\\
& Chat & & 3.49 & 119 & 0.88$\times$\\
& Avg. & & 4.56 & 155 & 1.15$\times$ \\
\bottomrule
\end{tabular}
\end{table}

\textbf{Acceptance results}\quad
Table~\ref{tab:dflash2} evaluates cross-target draft transfer, target-matched training, and draft precision across all six benchmarks with a maximum generation length of 256 tokens. Under the MNN-INT4
verification model, the mean accepted length of the MNN-trained
draft is lower than that of the MNN-INT4-trained draft by 1.06\% in
BF16 and 1.94\% in INT4. For the fully trained drafts evaluated
here, the mismatch between training source and verification model
incurs a modest acceptance penalty that remains acceptable
for deployment.

\textbf{On-device evaluation}\quad
Table~\ref{tab:dflash2-device} reports the on-device evaluation of the release configuration on AIME25, HumanEval, and MT-Bench, with a maximum generation length of 256 tokens. Speculative decoding achieves an average decode speedup of 1.65$\times$ on iPad Pro and 1.15$\times$ on Snapdragon 8 Gen 5, with math and code workloads benefiting more than chat. Despite similar mean accepted lengths across devices, the actual decode speedup varies with device, backend, and workload characteristics.

\section{Evaluation}
\label{sec:evaluation}

\subsection{Experimental Setup}

\textbf{Benchmarks}\quad
We evaluate the post-trained MNN-Pocket-0.3B on 11 benchmarks spanning Chinese and English knowledge, general reasoning, mathematics, coding, and instruction following: C-Eval~\cite{huang2023ceval}, CMMLU~\cite{cmmlu}, MMLU-Redux~\cite{gema2024mmlu-redux} and BBH~\cite{bbh} for knowledge; GSM8K~\cite{cobbe2021gsm8k}, MGSM-zh~\cite{mgsm}, and MATH-500~\cite{math500} for mathematics; HumanEval~\cite{humaneval} and MBPP~\cite{mbpp} for coding; and IFEval~\cite{ifeval} and IFEval-zh, our Chinese translation of IFEval, for instruction following. For the pre-trained MNN-Pocket-0.3B-Base, we additionally include HellaSwag~\cite{zellers2019hellaswag}, ARC-Easy~\cite{allenai:arc}, and GPQA-Diamond~\cite{gpqa-diamond} for general capabilities, and CRUX-O~\cite{gu2024cruxeval} for coding.



\begin{table}[!b]
  \centering
  \small
  \caption{Performance of post-trained MNN-Pocket-0.3B model and other baselines across 11 benchmarks.}
  \label{tab:main_results}
  \begingroup
  \definecolor{mainresultsblue}{HTML}{1D6FD0}
  \footnotesize
  \renewcommand{\arraystretch}{1.2}
  \setlength{\tabcolsep}{1pt}
  \renewcommand{\tabularxcolumn}[1]{m{#1}}
  \newcommand{\mainresultsmodel}[3][mainresultsblue]{%
    \parbox[c][5ex][c]{\linewidth}{\centering\fontsize{6.5}{8}\selectfont\color{#1}\mbox{#2}\\\mbox{#3}}%
  }
  \arrayrulecolor{mainresultsblue!25}
  \begin{tabularx}{\linewidth}{
    >{\raggedright\arraybackslash}m{0.14\linewidth}
    *{2}{>{\columncolor{mainresultsblue!8}\centering\arraybackslash}X}
    *{4}{>{\centering\arraybackslash}X}|
    *{3}{>{\color{black!45}\centering\arraybackslash}X}}
    \toprule
    & \multicolumn{6}{c}{\color{mainresultsblue}\bfseries $\sim$300M Models}
    & \multicolumn{3}{c}{\color{mainresultsblue!45}\bfseries $\sim$500M Models} \\
    \cline{2-10}
    {\color{mainresultsblue}\bfseries Benchmark}
    & \mainresultsmodel{MNN-Pocket}{0.3B}
    & \mainresultsmodel{MNN-Pocket}{0.3B-INT4}
    & \mainresultsmodel{Gemma-3}{270M}
    & \mainresultsmodel{ERNIE-4.5}{0.3B}
    & \mainresultsmodel{LFM2.5}{350M}
    & \mainresultsmodel{SmolLM2}{360M}
    & \mainresultsmodel[mainresultsblue!45]{Hunyuan}{0.5B}
    & \mainresultsmodel[mainresultsblue!45]{MiniCPM4}{0.5B}
    & \mainresultsmodel[mainresultsblue!45]{Qwen3}{0.6B} \\
    \midrule
    {\bfseries Average} & {\bfseries 50.00} & 49.59 & 6.42 & \underline{31.99} & 31.28 & 18.43 & 31.22 & 46.81 & 44.28 \\
    \hline
    \rowcolor{mainresultsblue!14}
    \multicolumn{10}{l}{\rule{0pt}{2.5ex}\color{mainresultsblue}\bfseries Knowledge} \\
    C-Eval & {\bfseries 38.86} & 40.27 & 6.84 & \underline{32.17} & 32.10 & 23.77 & 35.29 & 67.98 & 44.06 \\ \hline
    CMMLU & {\bfseries 40.11} & 39.65 & 4.94 & \underline{33.91} & 31.65 & 24.87 & 35.69 & 66.04 & 44.50 \\ \hline
    MMLU-Redux & {\bfseries 44.00} & 43.40 & 13.58 & 32.68 & \underline{41.01} & 24.17 & 36.29 & 52.85 & 51.63 \\ \hline
    BBH & {\bfseries 47.38} & 46.63 & 8.09 & \underline{31.07} & 26.03 & 27.74 & 54.65 & 47.61 & 41.94 \\ \hline
    \rowcolor{mainresultsblue!14}
    \multicolumn{10}{l}{\rule{0pt}{2.5ex}\color{mainresultsblue}\bfseries IF} \\
    IFEval & {\bfseries 73.20} & 74.12 & 20.15 & 44.92 & \underline{72.09} & 32.16 & 24.58 & 52.13 & 56.93 \\ \hline
    IFEval-zh & {\bfseries 74.59} & 72.49 & 17.02 & 32.87 & \underline{44.06} & 25.64 & 25.87 & 30.30 & 50.35 \\ \hline
    \rowcolor{mainresultsblue!14}
    \multicolumn{10}{l}{\rule{0pt}{2.5ex}\color{mainresultsblue}\bfseries Math} \\
    GSM8K & {\bfseries 54.36} & 52.84 & 0.00 & 31.24 & \underline{33.66} & 11.60 & 36.01 & 42.46 & 45.49 \\ \hline
    MGSM-zh & {\bfseries 51.20} & 48.80 & 0.00 & \underline{25.20} & 18.00 & 3.60 & 34.80 & 37.60 & 40.00 \\ \hline
    MATH500 & {\bfseries 35.40} & 35.40 & 0.00 & \underline{21.40} & 14.80 & 4.60 & 35.80 & 23.00 & 45.20 \\ \hline
    \rowcolor{mainresultsblue!14}
    \multicolumn{10}{l}{\rule{0pt}{2.5ex}\color{mainresultsblue}\bfseries Code} \\
    HumanEval & {\bfseries 53.66} & 53.66 & 0.00 & \underline{38.41} & 14.63 & 0.00 & 14.63 & 48.17 & 36.59 \\ \hline
    MBPP & {\bfseries 37.20} & 38.20 & 0.00 & \underline{28.00} & 16.00 & 24.60 & 9.80 & 46.80 & 30.40 \\
    \bottomrule
  \end{tabularx}
  \arrayrulecolor{black}
  \endgroup
\end{table}

\textbf{Baseline Models}\quad
We compare MNN-Pocket-0.3B with seven open-source language models: Gemma-3-270M~\citep{Gemma}, ERNIE-4.5-0.3B~\citep{ERNIE}, LFM2.5-350M~\citep{LFM}, SmolLM2-360M~\citep{SmolLM}, Hunyuan-0.5B~\citep{Hunyuan}, MiniCPM4-0.5B~\citep{MiniCPM}, and Qwen3-0.6B~\citep{Qwen3}.
These baselines span approximately 0.27B to 0.6B parameters, covering both models of comparable size and larger alternatives.
Table~\ref{tab:main_results} places MNN-Pocket-0.3B and its INT4 variant first, followed by baselines grouped into approximately 300M- and 500M-parameter classes.
For the base-model evaluation, we use the corresponding base or pre-trained versions of these models where available, as reported in Table~\ref{tab:base_results}.

\subsection{Post-trained Model Results}
We show the evaluation results of MNN-Pocket-0.3B and baseline models in Table~\ref{tab:main_results}. For knowledge assessment, we use the zero-shot chain-of-thought (CoT) variants of C-Eval, CMMLU, and MMLU-Redux. General reasoning is evaluated on BBH with 3-shot CoT prompting. Mathematical reasoning is evaluated on GSM8K and MATH500 with 4-shot CoT prompting, and on the Chinese subset of MGSM with 8-shot CoT prompting. For coding and instruction following, we use HumanEval, MBPP, IFEval, and IFEval-zh in the zero-shot setting.

MNN-Pocket-0.3B achieves the best average score of 50.00, outperforming the strongest ${\sim}300$M baseline, ERNIE-4.5-0.3B, by 18.0 points. Despite having 1.7--2$\times$ fewer parameters, it surpasses the larger $\sim$500M baselines, MiniCPM4-0.5B (46.81) and Qwen3-0.6B (44.28). The advantage is not concentrated on a single capability. Among the full-precision models, MNN-Pocket-0.3B ranks among the top three on all eleven benchmarks and achieves the best score on five of them, spanning instruction following (IFEval 73.20 and IFEval-zh 74.59), bilingual mathematical reasoning (GSM8K 54.36 and MGSM$_{\mathrm{zh}}$ 51.20), and code generation (HumanEval 53.66). The baselines are markedly more specialized: LFM2.5-350M matches MNN-Pocket-0.3B on English IFEval (72.09) but falls to 14.80 on MATH500 and 14.63 on HumanEval, MiniCPM4-0.5B is strong on knowledge-intensive tasks (C-Eval 67.98) yet reaches only 30.30 on IFEval-zh, and Hunyuan-0.5B leads on BBH (54.65) but scores 24.58 on IFEval. This balanced profile is consistent with our post-training design, in which the SFT Loop Engine (Section~\ref{sec:sft-loop-engine}) turns evaluation failures into targeted training data and multi-teacher on-policy distillation (Section~\ref{sec:mopd-teachers}) consolidates four domain specialists into a single student.

\subsection{Pre-trained Model Results}

\begin{table}[!b]
  \centering
  \small
  \caption{Performance of pre-trained MNN-Pocket-0.3B-Base model and other baselines across 11 benchmarks.}
  \label{tab:base_results}
  \begingroup
  \definecolor{mainresultsblue}{HTML}{1D6FD0}
  \footnotesize
  \renewcommand{\arraystretch}{1.2}
  \setlength{\tabcolsep}{1pt}
  \renewcommand{\tabularxcolumn}[1]{m{#1}}
  \newcommand{\mainresultsmodel}[3][mainresultsblue]{%
    \parbox[c][5ex][c]{\linewidth}{\centering\fontsize{6.5}{8}\selectfont\color{#1}\mbox{#2}\\\mbox{#3}}%
  }
  \arrayrulecolor{mainresultsblue!25}
  \begin{tabularx}{\linewidth}{
    >{\raggedright\arraybackslash}m{0.14\linewidth}
    >{\columncolor{mainresultsblue!8}\centering\arraybackslash}X
    *{4}{>{\centering\arraybackslash}X}|
    *{2}{>{\color{black!45}\centering\arraybackslash}X}}
    \toprule
    & \multicolumn{5}{c}{\color{mainresultsblue}\bfseries $\sim$300M Models}
    & \multicolumn{2}{c}{\color{mainresultsblue!45}\bfseries $\sim$500M Models} \\
    \cline{2-8}
    {\color{mainresultsblue}\bfseries Benchmark}
    & \mainresultsmodel{MNN-Pocket}{0.3B-Base}
    & \mainresultsmodel{Gemma-3}{270M Base}
    & \mainresultsmodel{ERNIE-4.5}{0.3B Base}
    & \mainresultsmodel{LFM2.5}{350M Base}
    & \mainresultsmodel{SmolLM2}{360M Base}
    & \mainresultsmodel[mainresultsblue!45]{Hunyuan}{0.5B-Pretrain}
    & \mainresultsmodel[mainresultsblue!45]{Qwen3}{0.6B Base} \\
    \midrule
    {\bfseries Average} & \underline{33.58} & 16.94 & {\bfseries 34.00} & 29.86 & 22.25 & 42.60 & 44.86 \\
    \hline
    \rowcolor{mainresultsblue!14}
    \multicolumn{8}{l}{\rule{0pt}{2.5ex}\color{mainresultsblue}\bfseries General Tasks} \\
    C-Eval & \underline{36.55} & 23.25 & {\bfseries 40.56} & 35.66 & 22.07 & 57.95 & 54.68 \\ \hline
    CMMLU & \underline{38.13} & 25.26 & {\bfseries 39.54} & 36.78 & 25.30 & 54.54 & 53.29 \\ \hline
    MMLU-Redux & \underline{47.07} & 24.47 & 46.59 & {\bfseries 48.44} & 30.19 & 59.34 & 57.45 \\ \hline
    HellaSwag & \underline{38.25} & 29.29 & 36.33 & 36.65 & {\bfseries  43.03} & 36.87 & 40.98 \\
    ARC-Easy & 65.82 & 57.70 & 65.03 & \underline{70.16} & {\bfseries 70.59} & 66.75 & 65.57 \\
    GPQA-Diamond & {\bfseries 27.27} & \underline{25.76} & 24.24 & 21.72 & 21.72 & 25.25 & 30.30 \\
    \hline
    \rowcolor{mainresultsblue!14}
    \multicolumn{8}{l}{\rule{0pt}{2.5ex}\color{mainresultsblue}\bfseries Math} \\
    GSM8K & \underline{26.00} & 0.00 & {\bfseries 31.92} & 19.79 & 4.55 & 42.99 & 58.38 \\ \hline
    MGSM-zh & \underline{17.60} & 0.00 & {\bfseries 21.60} & 16.00 & 1.60 & 45.20 & 48.40 \\ \hline
    MATH500 & \underline{16.80} & 0.00 & {\bfseries 17.20} & 16.20 & 5.20 & 23.00 & 29.20 \\ \hline
    \rowcolor{mainresultsblue!14}
    \multicolumn{8}{l}{\rule{0pt}{2.5ex}\color{mainresultsblue}\bfseries Code} \\
    HumanEval & {\bfseries 28.66} & 0.61 & \underline{27.44} & 11.59 & 0.00 & 28.05 & 28.66 \\ \hline
    CRUX-O & {\bfseries 27.27} & 0.00 & \underline{23.50} & 15.50 & 20.50 & 28.63 & 26.50 \\
    \bottomrule
  \end{tabularx}
  \arrayrulecolor{black}
  \endgroup
\end{table}
We show the evaluation results of MNN-Pocket-0.3B-Base and baseline models in Table~\ref{tab:base_results}. C-Eval, CMMLU, MMLU-Redux, and GPQA-Diamond are evaluated in the 5-shot setting, HellaSwag in the 10-shot setting, and ARC-Easy in the zero-shot setting. For mathematics, we use 4-shot prompting on GSM8K and MATH500 and 8-shot prompting on MGSM-zh. HumanEval and CRUX-O are evaluated in the zero-shot setting.

MNN-Pocket-0.3B-Base achieves an average score of 33.58, ranking second among the approximately 300M models and trailing ERNIE-4.5-0.3B Base (34.00) by only 0.42 points. It exceeds LFM2.5-350M Base, SmolLM2-360M Base, and Gemma-3-270M Base by 3.72, 11.33, and 16.64 points, respectively. Within this size group, it ranks among the top two on ten of the eleven benchmarks and leads on GPQA-Diamond (27.27), HumanEval (28.66), and CRUX-O (27.27), showing competitive performance across both general and specialized tasks. The larger Qwen3-0.6B Base and Hunyuan-0.5B-Pretrain models retain higher overall averages of 44.86 and 42.60, respectively, with stronger knowledge and mathematical reasoning results. Nevertheless, MNN-Pocket-0.3B-Base matches Qwen3-0.6B Base on HumanEval (28.66) and exceeds it on CRUX-O (27.27 vs.\ 26.50). These results establish a competitive pre-training foundation at the 300M scale, while leaving room for further gains in knowledge and mathematics.

\subsection{MOPD Training Gains}
\begin{figure*}[!b]
    \centering
    \includegraphics[width=\textwidth]{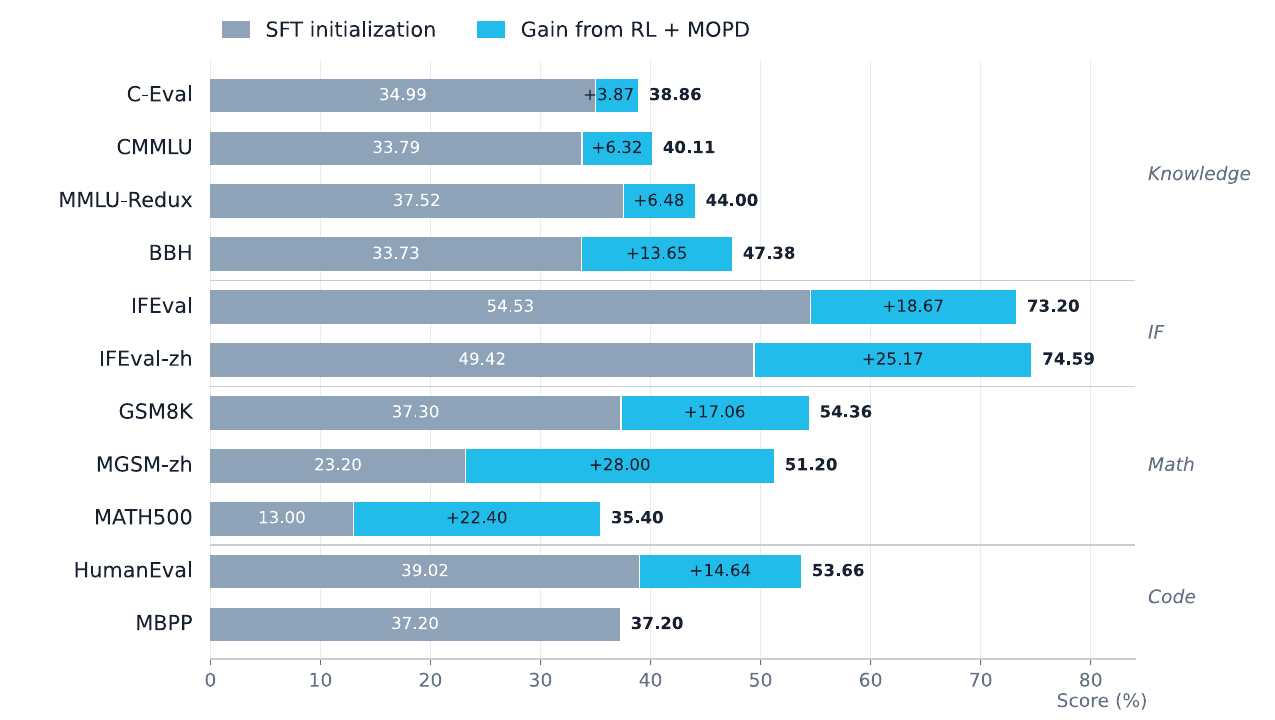}
    \caption{
        Benchmark evolution from the SFT initialization to the final MOPD model.
        Gray-blue bars denote SFT scores, and blue segments denote gains after MOPD.
        Final scores match Table~\ref{tab:main_results}.
    }
    \label{fig:sft_to_mopd_evolution}
\end{figure*}

Figure~\ref{fig:sft_to_mopd_evolution} compares the final MOPD model against its SFT initialization on the same 11 benchmarks as Table~\ref{tab:main_results}. MOPD improves ten tasks and maintains the MBPP score at 37.20, raising the macro-average from 35.79 to 50.00. The largest gains occur on MGSM$_{\mathrm{zh}}$ (+28.00), IFEval-zh (+25.17), and MATH500 (+22.40). These results show improvements across knowledge, instruction following, mathematics, and code generation.

MNN-Pocket-0.3B-INT4 achieves an equal-weight average of 49.59 across the same 11 tasks, only 0.41 points below its BF16 counterpart (50.00), indicating that our quantization approach largely preserves floating-point performance. The Self-OPD and quantization-aware training procedure is described in Section~\ref{sec:quantization-aware-training}.

Overall, the results validate the post-training pipeline: supervised fine-tuning with the SFT Loop Engine and capability consolidation produce a single 0.3B student whose abilities are broad rather than specialized, while the deployment path keeps the teachers out of inference.





\section{Applications}

To assess the practical value of MNN-Pocket-0.3B beyond benchmark scores, we evaluate it in two realistic workflows: (1) a structured E-Commerce Review Opinion Mining task requiring joint span extraction and fine-grained classification under strict format constraints, and (2) a fully on-device, streaming Voice Assistant deployed on mobile hardware to evaluate real-time conversational latency and interactive quality.

\subsection{MNN-Pocket-0.3B for E-Commerce Review Opinion Mining}

To assess the semantic understanding of the model trained in this work under a controlled, externally scored setting, we evaluate it on the E-Commerce Review Opinion Mining track of the Tianchi Classic Leaderboard.\footnote{\url{https://tianchi.aliyun.com/competition/entrance/532421/information}} The corpus consists of Chinese cosmetics reviews in which brand mentions are masked with \texttt{**} for anonymity, so brand identity is unavailable as a cue. Given a review, a system must extract every opinion expressed in it as a quadruple (AspectTerms, OpinionTerms, Categories, Polarities). The aspect and opinion terms must be reproduced verbatim from the review, the category is drawn from thirteen predefined classes, and the polarity is one of positive, neutral, and negative. The annotation guidelines allow either term to be absent, in which case the field is filled with the sentinel \texttt{\_}, so a system must decide both what is being evaluated and whether it is stated explicitly at all.

\begin{table}[!b]
\small
\caption{Prompt and example for E-Commerce Review Opinion Mining. }
\label{tab:ecommerce-task}
\begin{tabular}{@{}p{0.12\linewidth}p{0.82\linewidth}@{}}
\toprule
System & 你是一个聪明的AI电商助手。 \\
\midrule
Input  & 请在商品评论中抽取：\newline
         1.评论中被评价的商品属性（AspectTerms），并为其归入对应类目（Categories）。可选类目为：['包装', '成分', '尺寸', '服务', '功效', '价格', '气味', '使用体验', '物流', '新鲜度', '真伪', '整体', '其他']；\newline
         2.消费者用以评价的表述（OpinionTerms），并判定其情感倾向（Polarities）。可选倾向为：['正面', '负面', '中性']。\newline
         若评论未明确提及被评价的属性，AspectTerms 置为空字符串 ''。\newline
         供参考的评论示例：'很好，遮暇功能差一些'\newline
         输出：[\{'AspectTerms': '', 'OpinionTerms': '很好', 'Categories': '整体', 'Polarities': '正面'\}, \{'AspectTerms': '遮暇功能', 'OpinionTerms': '差一些', 'Categories': '功效', 'Polarities': '负面'\}]\newline
         下面是需要处理的评论：包装太随便了，感觉很不好\newline
         输出： \\
\midrule
Output & [\{'AspectTerms': '包装', 'OpinionTerms': '太随便了', 'Categories': '包装', 'Polarities': '负面'\}, \{'AspectTerms': '', 'OpinionTerms': '很不好', 'Categories': '整体', 'Polarities': '负面'\}] \\
\bottomrule
\end{tabular}
\end{table}

Table~\ref{tab:ecommerce-task} shows the prompt and an example instance. The system message fixes the assistant persona, the user turn states the extraction task, enumerates the admissible categories and polarities, provides one worked example, and finally supplies the review to be processed; the target is the quadruple list. The first quadruple names an aspect explicitly and falls in the packaging category, whereas the second has no aspect term: the field is left empty in the target and is written back as the sentinel \texttt{\_} at submission time, yet the opinion must still be assigned a category --- here the overall class --- and a polarity. Note also that one review routinely yields several quadruples, all of which have to be recovered.

We adopt this task as a probe of semantic understanding for three reasons. First, it couples span-level grounding with fine-grained classification: a prediction is credited only when all four fields are simultaneously correct, so partial pattern matching on sentiment-bearing words does not score. Second, polarity in this corpus is frequently carried by mild or concessive expressions rather than explicit sentiment lexicon, which penalises models that rely on surface cues. Third, the output is a strictly formatted structured list, so the task jointly measures instruction following and format stability under a fixed decoding budget. Because all submissions are scored by the organiser against held-out labels, the resulting comparison across language backbones is external and reproducible.

\subsubsection{Dataset and Training Setup}
\label{sec:review_training}

The training split contains 3,229 reviews annotated with 6,633 quadruples, an average of 2.05 opinions per review. The test split contains 2,237 reviews whose labels are withheld; scores are obtained by submitting predictions to the leaderboard. The thirteen categories are packaging, ingredients, size, service, efficacy, price, scent, user experience, logistics, freshness, authenticity, overall, and other, and the three polarities are positive, neutral, and negative; both label sets are fixed by the training split. The official annotation additionally records the character offsets of every aspect and opinion span, but the evaluation ignores position information and scores the quadruples alone, so we discard the offsets and supervise the quadruple list only. We convert the official annotation tables into a single-turn supervised fine-tuning corpus of 3,229 examples. Each user turn states the extraction task, enumerates the thirteen categories and the three polarities, and provides one worked example; the target is the gold quadruple list rendered as a Python literal. All backbones share the same system prompt and the same instruction template, and no development split is held out. At submission time the generated list is flattened into the official \texttt{id, AspectTerms, OpinionTerms, Categories, Polarities} format, empty fields are written back as the sentinel \texttt{\_}, and reviews for which no quadruple is produced receive a placeholder row so that every identifier appears exactly once.


All backbones are fine-tuned with LoRA-based supervised fine-tuning under ms-swift~\cite{zhao2025swiftascalablelightweightinfrastructure} in bfloat16 on a single AMD accelerator. We train for three epochs with a learning rate of $1\times10^{-4}$, a warmup ratio of 0.05, a maximum sequence length of 4,096, and a per-device batch size of 16 without gradient accumulation. Data, prompt template, system prompt, and every optimisation hyperparameter are held identical across backbones\footnote{Gemma-3's chat template does not accept a system role; for that backbone the same system message is prepended to the first user turn instead, so the text the model sees is unchanged.}, so the differences reported below are attributable to the pretrained model rather than to per-model tuning. For backbones with an optional reasoning mode we disable it, matching the training distribution.

\subsubsection{Evaluation}

Table~\ref{tab:review_mining_results} reports F1 on the official test split. A predicted quadruple is credited only when all four fields match a gold quadruple exactly, and F1 is micro-averaged over the split.
\begin{table}[!b]
\centering
\caption{\textbf{E-Commerce Review Opinion Mining results.} All backbones are LoRA fine-tuned with the identical recipe of Section~\ref{sec:review_training} and scored on the official test split; a quadruple counts as correct only under exact match on all four fields. Parameter counts are the nominal sizes given by the model names. Rows are ordered by F1.}
\label{tab:review_mining_results}
\small
\setlength{\tabcolsep}{6pt}
\begin{tabular}{lcc}
\toprule
Language backbone        & Params & F1 (\%) $\uparrow$ \\
\midrule
\textbf{MNN-Pocket-0.3B} & 0.3B   & \textbf{80.15}     \\
Qwen3-1.7B               & 1.7B   & 78.59              \\
Qwen3-0.6B               & 0.6B   & 77.47              \\
Hunyuan-0.5B             & 0.5B   & 77.27              \\
ERNIE-4.5-0.3B           & 0.3B   & 76.73              \\
LFM2.5-350M              & 0.35B  & 76.19              \\
MiniCPM4-0.5B            & 0.5B   & 75.27              \\
LFM2.5-230M              & 0.23B  & 73.14              \\
Gemma-3-270M             & 0.27B  & 72.23              \\
SmolLM2-360M             & 0.36B  & 67.54              \\
\bottomrule
\end{tabular}
\end{table}

MNN-Pocket-0.3B leads at 80.15, ahead of the runner-up Qwen3-1.7B (78.59) by 1.56 points with less than a fifth of its parameters, and of Qwen3-0.6B (77.47) by 2.68 points with half. It also beats both 0.5B systems (Hunyuan-0.5B by 2.88 points and MiniCPM4-0.5B by 4.88) and the identically sized ERNIE-4.5-0.3B by 3.42, making it both the strongest and the most parameter-efficient point in the comparison --- the regime that matters for on-device deployment. The field spans 12.61 points and its ordering does not track capacity: the weakest system, SmolLM2-360M, is not the smallest, LFM2.5-350M outscores MiniCPM4-0.5B, and a 0.3B backbone beats every larger one, indicating that once in-domain supervision is available the binding constraint is adaptability rather than capacity. Since a quadruple is credited only when its aspect span, opinion span, category, and polarity are all correct at once, MNN-Pocket-0.3B's clear lead shows that it sustains span-level grounding, fine-grained classification, and stable-format instruction following together --- at less than a fifth of the parameters of the next-best model, precisely the property on-device deployment requires.

\subsection{An Efficient On-Device Voice Assistant with MNN-Pocket-0.3B}

To assess the deployment feasibility and real-time interaction capability of the model trained in this work, we built a fully on-device speech-to-speech dialogue system. The system uses MNN-Pocket-0.3B as its language backbone and integrates streaming automatic speech recognition (ASR), voice activity detection (VAD), semantic turn-end detection, streaming text generation, incremental text-to-speech (TTS) synthesis, and barge-in support. All inference and audio processing are performed locally, eliminating network-induced latency and dependence on cloud inference services while keeping user speech and dialogue data on the device.

\begin{figure}[!b]
    \centering
    \includegraphics[width=1.0\linewidth]{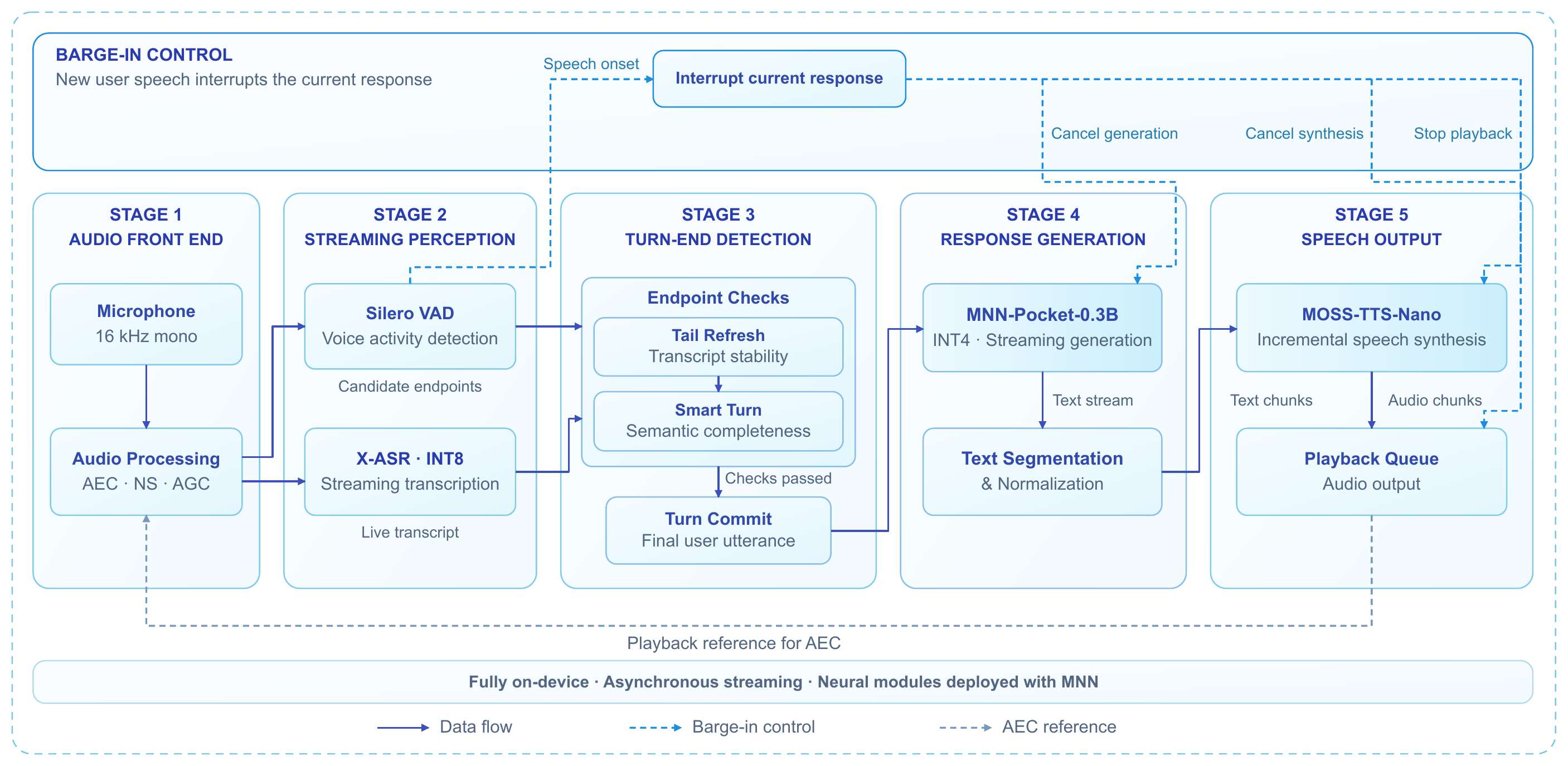}
    \caption{\textbf{Fully on-device speech-to-speech dialogue pipeline.} Processed microphone audio feeds Silero VAD and INT8 X-ASR for speech activity detection and streaming transcription. Transcript-stability checks and Smart Turn determine when to commit the user utterance to INT4 MNN-Pocket-0.3B. Generated text is segmented and normalized for incremental synthesis by MOSS-TTS-Nano, with audio chunks queued for playback. Generation, synthesis, and playback run asynchronously; new user speech interrupts all three, while the playback signal provides a reference for acoustic echo cancellation. All neural modules run locally with MNN.}
    \label{fig:application_1}
\end{figure}

\subsubsection{System Pipeline}

Figure~\ref{fig:application_1} shows the overall system pipeline. All neural-network modules are deployed using the MNN inference framework. Speech recognition is performed by an INT8-quantized X-ASR~\cite{xasr} model, text responses are generated by the INT4-quantized MNN-Pocket-0.3B model trained in this work, and speech is synthesized by MOSS-TTS-Nano~\cite{mosstts}.

User speech is captured by the microphone and processed by the audio front end for acoustic echo cancellation, noise suppression, and automatic gain control. The processed audio is then fed to Silero VAD~\cite{SileroVAD} for speech activity detection and X-ASR for streaming transcription. Turn-end detection follows a two-stage strategy. Silero VAD first identifies candidate endpoints from speech activity, after which Smart Turn~\cite{smartturnv3} evaluates whether the current utterance is semantically complete. Combined with transcript-stability checks, this strategy reduces the risk of premature turn termination that may arise when endpoint detection relies solely on silence duration.

Once the user turn is committed, the finalized utterance is passed to the MNN-Pocket-0.3B model for streaming text generation. The generated text is incrementally segmented and normalized into text chunks, which are then fed to MOSS-TTS-Nano for incremental speech synthesis. The resulting audio chunks are placed in a playback queue for sequential output. Text generation, speech synthesis, and audio playback operate asynchronously, allowing playback to begin before the full response has been generated.

During playback, the system continues to monitor incoming speech. When new user speech is detected, it cancels ongoing text generation and speech synthesis and stops audio playback to support barge-in. The playback signal is also fed back to the audio front end as a reference for acoustic echo cancellation.

\begin{table*}[!b]
\centering
\caption{On-device performance of the MNN-Pocket-0.3B Voice Assistant.}
\label{tab:on_device_performance}
\small
\setlength{\tabcolsep}{12pt}
\renewcommand{\arraystretch}{1.12}
\begin{tabular}{@{}lccc@{}}
\toprule
Metric & iPhone & iPad & Android \\
\midrule
Device / SoC
& 13 Pro / A15 & Pro / M5 & Canoe / 8Gen5 \\

\addlinespace[4pt]
Model storage (MB) $\downarrow$
& 2,018 & 2,018 & 2,018 \\
Peak memory (MB) $\downarrow$
& 1,591 & 1,858 & 1,704 \\
ASR RTF $\downarrow$
& 0.073 & 0.047 & 0.045 \\
LLM decoding speed (token/s) $\uparrow$
& 130.2 & 324.9 & 231.7 \\
TTS RTF $\downarrow$
& 0.653 & 0.366 & 0.651 \\

\addlinespace[4pt]
Theoretical TTFT (ms) $\downarrow$
& $\sim$169 & $\sim$\textbf{46} & $\sim$144 \\
TTFT P50/P95 (ms) $\downarrow$
& 568 / 585 & \textbf{472 / 481} & 523 / 535 \\

\addlinespace[2pt]
Theoretical TTFA (ms) $\downarrow$
& $\sim$485 & $\sim$\textbf{171} & $\sim$358 \\
TTFA P50/P95 (ms) $\downarrow$
& 1,055 / 1,113 & \textbf{600 / 618} & 935 / 1,022 \\

\bottomrule
\end{tabular}
\end{table*}

\begin{table*}[!b]
\centering
\caption{VoiceBench results of the MNN-Pocket-0.3B Voice Assistant and other models.}
\label{tab:voicebench_results}
\resizebox{\textwidth}{!}{
\begin{tabular}{l*{9}{c}@{\hspace{1.2em}}c}
\toprule
Language backbone
& Common
& WildVoice
& SD-QA
& OBQA
& MMSU
& BBH
& IFEval
& AdvBench
& AlpacaEval
& \textbf{Overall} \\
\midrule
\textbf{MNN-Pocket-0.3B}
& 40.23 & 36.72 & 5.75 & \underline{32.75} & 27.91
& \textbf{58.90} & \textbf{64.11}
& \textbf{97.69} & 44.91 & \textbf{45.44} \\

Gemma3-270M
& 37.17 & 31.54 & 6.43 & 25.93 & 25.02
& 49.40 & 18.32 & \underline{93.27} & 38.56 & 36.18 \\

ERNIE-4.5-0.3B
& 37.33 & 32.50 & 4.87 & 28.57 & 25.63
& \underline{50.60} & 33.98 & 39.42 & 38.46 & 32.37 \\

LFM2.5-350M
& \textbf{47.17} & \textbf{39.96} & 11.06 & 22.86 & 25.76
& 46.30 & \underline{57.17} & 88.65
& \textbf{50.47} & \underline{43.27} \\

SmolLM2-360M
& 37.93 & 32.26 & \textbf{15.78} & 21.98 & 24.59
& 46.40 & 32.59 & 23.46 & 41.20 & 30.69 \\

Hunyuan-0.5B
& 22.67 & 22.19 & 2.02 & 29.67 & 28.24
& 48.70 & 18.06 & 23.85 & 23.73 & 24.35 \\

MiniCPM4-0.5B
& 43.90 & 35.69 & 10.67 & 28.79 & \underline{31.13}
& 49.70 & 44.04 & 58.65 & 43.76 & 38.48 \\

Qwen3-0.6B
& \underline{45.87} & \underline{38.77} & \underline{11.15}
& \textbf{35.82} & \textbf{32.76}
& 49.60 & 47.73 & 60.96 & \underline{47.70} & 41.15 \\
\bottomrule
\end{tabular}
}
\end{table*}

\subsubsection{Evaluation}

To assess the on-device efficiency, inference speed, and runtime latency profile of the proposed voice assistant, we conduct comprehensive evaluations across various mobile devices. We report both the CPU-based compute-only theoretical latency and the measured production latency. The theoretical latency assumes warmed-up models, immediate endpoint commitment, and synthesis from the earliest available text fragment. The production configuration introduces additional safeguards for ASR tail completeness, semantic turn detection, and TTS prosody stability. Table~\ref{tab:on_device_performance} summarizes the overall performance. The theoretical latency estimates exclude endpointing delays and text-buffering safeguards, whereas measured latencies are reported as the 50th and 95th percentiles under warmed-up conditions.

To comprehensively assess the proposed system's spoken dialogue capability and intelligent conversational performance on standard voice benchmarks, we evaluate all language backbones using a unified setup with an X-ASR front end (160-ms chunk size and 2-s trailing silence). The system takes speech as input and simultaneously produces text and speech responses. Open-ended tasks are scored by Qwen3.7-Max with three votes per sample. Table~\ref{tab:voicebench_results} reports results on nine VoiceBench~\cite{chen2026voicebench} tasks. MNN-Pocket-0.3B achieves the highest overall score of 45.44, surpassing the second-best backbone, LFM2.5-350M, by 2.17 points. It ranks first on BBH, IFEval, and AdvBench, demonstrating competitive reasoning, instruction-following, and safety performance among the evaluated backbones.

\section{Conclusion and Limitations}

We present MNN-Pocket-0.3B, an on-device language model that combines architecture–tokenizer co-design, staged training, and MNN runtime optimization to balance capability, memory, and latency. It delivers strong benchmark performance and supports practical opinion mining and streaming voice interaction. 

However, knowledge-intensive tasks and complex reasoning remain challenging, and deployment efficiency varies across devices and workloads. Our evaluations focus primarily on Chinese and English and a limited set of applications and hardware. Future work will strengthen model reliability, broaden evaluation coverage, improve end-to-end efficiency, and explore on-device agents, search and recommendation, user intent understanding, and other e-commerce applications within the Taobao mobile app.


\section{Contributions and Acknowledgments}

MNN-Pocket-0.3B represents MNN’s exploration of on-device LLM training and deployment and is the result of the sustained, collective efforts of all our team members and interns. This work was also supported by the Key Research and Development Program of Zhejiang Province (Grant No. 2025C01026). \

\textbf{Project Leaders} Zhiwen Chen, Zhaode Wang, Chengfei Lv, Gang Yu

\textbf{Contributors} (Ordered by the last name)
Juncan Deng, Haotian Dong, Zhewei Fang, Canan Huang, Zhoulin Ji, Zefeng Sun, Gexing Tan, Yunfeng Wang, Yufan Xiong, Huaxiang Zhang, Weijun Zhang, Yuxin Zhang

\newpage

\clearpage

\bibliographystyle{plainnat}
\nocite{hendrycks2021math}
\bibliography{main}


\end{CJK*}
\end{document}